\PassOptionsToPackage{unicode}{hyperref}
\PassOptionsToPackage{hyphens}{url}
\PassOptionsToPackage{dvipsnames,svgnames*,x11names*,table}{xcolor}

\documentclass[11pt,
    a4paper,
    american,
    numbers=noenddot, % not 1.1. but 1.1 
    oneside,
    bibliography=totoc,
    bibtotocnumbered,
    listof=totoc,
    parskip=half,
]{scrreprt}

\usepackage{geometry}

\usepackage{multirow}
\usepackage{hhline}
\usepackage{pdfpages}
\usepackage{comment}
\usepackage{amsmath}
\usepackage{amssymb}
\usepackage{stmaryrd}
\usepackage{xspace}

\usepackage[l2tabu]{nag}

\usepackage[utf8]{inputenc}
\usepackage[T1]{fontenc}

\usepackage{helvet}
\usepackage{lmodern}

\usepackage{xcolor}

\usepackage{graphicx}
\graphicspath{{images/}}

\usepackage{amsfonts}
\usepackage{amsmath}
\usepackage{mathabx}
\usepackage{MnSymbol}

\usepackage{mathtools}

\usepackage{varioref}
\usepackage[pdfencoding=unicode, psdextra]{hyperref}
\usepackage{cleveref}

\usepackage{zref}

\usepackage{lipsum}

\usepackage{setspace}

\usepackage{tabularx}
\usepackage{booktabs} % for \midrule
\usepackage{colortbl} % in conjunction with color this creates colored tables

\usepackage{paralist}

\usepackage{scrlayer-scrpage}

\KOMAoptions{
    headsepline = true,
    footsepline = true,
    plainfootsepline = true,
}

\automark[chapter]{chapter}

\clearpairofpagestyles

\ohead{\headmark}
\ihead{\author}
\ofoot*{\pagemark}

\setkomafont{pageheadfoot}{\sffamily}
\setkomafont{pagination}{\sffamily}

\usepackage{tikz}
\usetikzlibrary{calc,positioning}

\usepackage{pgfplots}

\pgfplotsset{compat=newest}

\usepackage{adjustbox}

\usepackage{xparse}

\usepackage{picture}

\usepackage{siunitx}
\usepackage{ifthen}

\usepackage{biblatex}

\newcommand{\ten}[1]{ \ifthenelse{ \equal{A}{#1} \or \equal{B}{#1} \or
        \equal{C}{#1} \or \equal{D}{#1} \or \equal{E}{#1} \or \equal{C}{#1} \or
        \equal{D}{#1} \or \equal{E}{#1} \or \equal{F}{#1} \or \equal{G}{#1} \or
        \equal{H}{#1} \or \equal{I}{#1} \or \equal{J}{#1} \or \equal{K}{#1} \or
        \equal{L}{#1} \or \equal{M}{#1} \or \equal{N}{#1} \or \equal{P}{#1} \or
        \equal{Q}{#1} \or \equal{R}{#1} \or \equal{S}{#1} \or \equal{T}{#1} \or
        \equal{U}{#1} \or \equal{V}{#1} \or \equal{W}{#1} \or \equal{X}{#1} \or
        \equal{Y}{#1} \or \equal{Z}{#1} }{\ensuremath{\text{$\boldsymbol{#1}$}}}{
        \ensuremath{\text{\Large $\boldsymbol{#1}$}}}}

\usepackage{scrwfile}
\usepackage[toc,nomain,acronym,automake,numberedsection]{glossaries-extra}
\usepackage[toc,nomain,acronym,automake,numberedsection]{glossaries}

\newglossary[sym]{symbolslist}{sys}{syo}{Symbols}

\newglossary*{glossaryx}{Glossary}

\glsaddkey{unit}{}{\glsentryunit}{\Glsentryunit}{\glsunit}{\Glsunit}{\GLSunit}

\newglossarystyle{abkstyle}{%
    \renewenvironment{theglossary}%
    {\begin{longtable}{@{}p{.25\textwidth}@{}p{.75\textwidth}@{}}
            \endfirsthead
            \endhead
            \endfoot
            \endlastfoot}{\end{longtable}}%
    \renewcommand*{\glsgroupheading}[1]{\ifstrequal{A}{##1}{\relax}{\rlap{\large\textbf{\glsgetgrouptitle{##1}}} & \tabularnewline\addlinespace}}%
    }

\newglossarystyle{glossaryxstyle}{%
    \renewenvironment{theglossary}%
    {\begin{longtable}{@{}p{.4\textwidth}@{}p{.64\textwidth}@{}}
            \endfirsthead
            \endhead
            \endfoot
            \endlastfoot}{\end{longtable}}%
    \renewcommand*{\glsgroupheading}[1]{\ifstrequal{A}{##1}{\relax}{\rlap{\large\textbf{\glsgetgrouptitle{##1}}} & \tabularnewline\addlinespace}}%
    }

\newglossarystyle{symbunitlong}{%
    \setglossarystyle{long3col}% base this style on the list style
            {\end{longtable}}%

}

\makeglossaries

\definecolor{fh-mint}{RGB}{0,177,172}
\def\hlmint{\fontfamily{lmss}\fontsize{19pt}{19pt}\fontseries{b}\color{fh-mint}\selectfont}
\def\hlAbold{\fontfamily{lmss}\fontsize{14pt}{30pt}\fontseries{b}\selectfont}
\def\hlAnormal{\fontfamily{lmss}\fontsize{14pt}{14pt}\fontseries{n}\selectfont}

\newcommand{\showlogo}{
    \begin{picture}(0pt,0pt)(2cm,-2cm)
        \put(17cm,-4cm){\includegraphics[height=5cm]{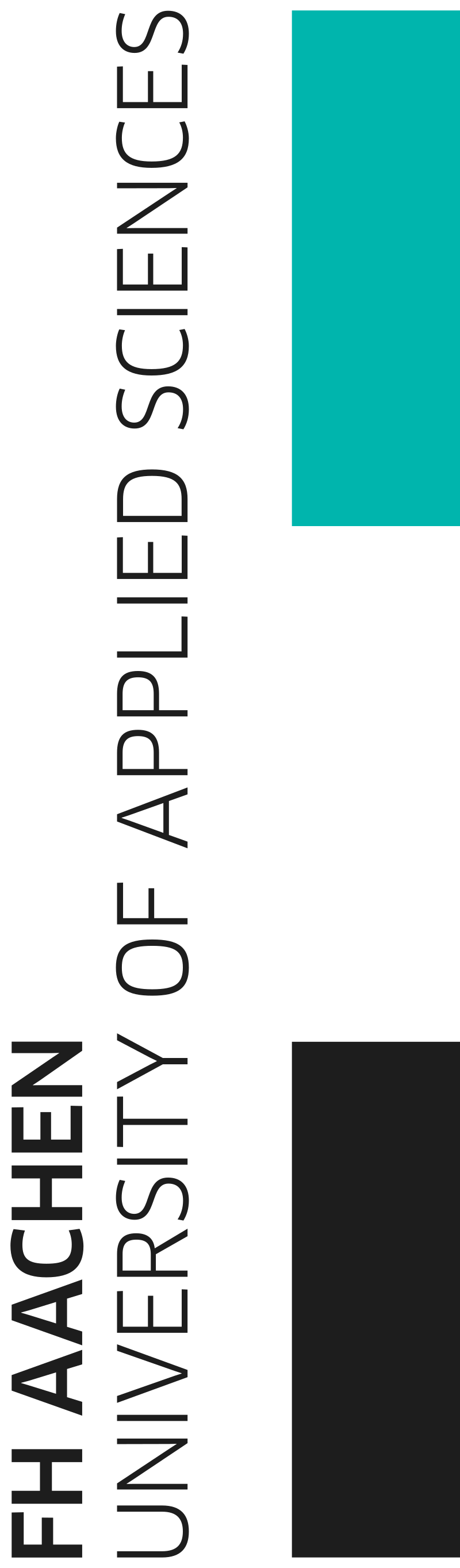}}
        \put(-0.5cm,-2.3cm){\includegraphics[height=5cm]{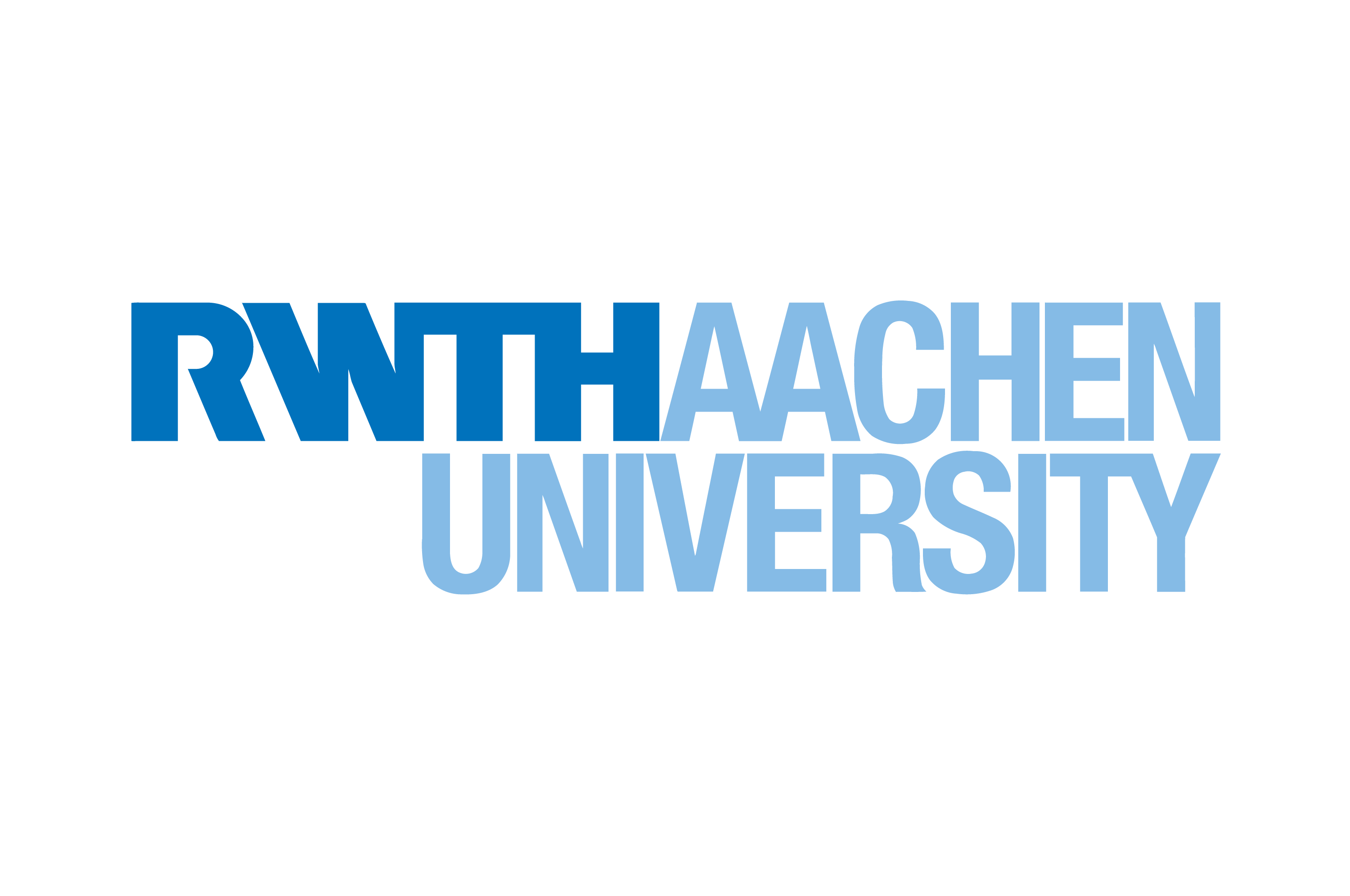}}
    \end{picture}
    
}

\ExplSyntaxOn
\keys_define:nn { titlepage }
{
    organization .tl_set:N  = \l__titlepage_organization_tl,
    title  .tl_set:N  = \l__titlepage_title_tl,
    type .tl_set:N  = \l__titlepage_type_tl,
    faculty .tl_set:N  = \l__titlepage_faculty_tl,
    department .tl_set:N  = \l__titlepage_department_tl,
    author  .tl_set:N  = \l__titlepage_author_tl,
    matrikel  .tl_set:N  = \l__titlepage_matrikel_tl,
    place  .tl_set:N  = \l__titlepage_place_tl,
    date .tl_set:N  = \l__titlepage_date_tl,
    
    examiner .tl_set:N  = \l__titlepage_examiner_tl,
    secondExaminer .tl_set:N  = \l__titlepage_second_examiner_tl,
    advisorFirst .tl_set:N  = \l__titlepage_advisorFirst_tl,
    advisorSecond .tl_set:N  = \l__titlepage_advisorSecond_tl,
}

\NewDocumentCommand { \makeTheTitlepage } { +m }
{
    \group_begin:
    \keys_set:nn { titlepage } { #1 }
    \title{\l__titlepage_title_tl}
    \author{\l__titlepage_author_tl}
    \date{\l__titlepage_date_tl}
    \begin{titlepage}
        \showlogo
        \vspace*{2cm}
        \begin{center}
            {
                {\hlmint \l__titlepage_organization_tl}

                \par \hlAbold
                \l__titlepage_faculty_tl

                \vspace*{0.5cm}

                \par
                \begin{spacing}{1.5}
                    \hlAnormal \l__titlepage_department_tl
                \end{spacing}
            }

            \vspace*{1cm}
            \par \l__titlepage_type_tl
            \par \hlAbold \l__titlepage_title_tl

            \vspace*{1cm}

            \par \hlAnormal \l__titlepage_author_tl
            \vspace*{0.5cm}
            \par \hlAnormal \l__titlepage_matrikel_tl
            \vspace*{0.5cm}
            \par \hlAnormal \l__titlepage_place_tl,~\l__titlepage_date_tl
            
            \vspace*{1cm}
            \par
            \begin{tabularx}{\textwidth}{ X  X  }
                Referent:        & \l__titlepage_examiner_tl        \\[2em]
                Koreferent:        & \l__titlepage_second_examiner_tl \\[2em]
                Forschungsleiter: &
                \l__titlepage_advisorFirst_tl \\[2em]
                Forschungsleiter: &
                \l__titlepage_advisorSecond_tl \\
            \end{tabularx}
            \par \hlAnormal
        \end{center}

    \end{titlepage}
    \group_end:
}
\ExplSyntaxOff

\begin{document}
\newabbreviation{ASIL}{ASIL}{Automotive Safety Integrity Level}
\newabbreviation{C}{C}{Controllability}
\newabbreviation{CEN}{CEN}{European Committee for Standardization (French: Comité Européen de Normalisation)}
\newabbreviation{CENELEC}{CENELEC}{European Committee for Electrotechnical Standards (fr.: Comité Européen de Normalisation Électrotechnique)}
\newabbreviation{DIN}{DIN}{German Institute for Standardization (German: Deutsches Institut für Normung)}
\newabbreviation{E}{E}{Exposure}
\newabbreviation{AELV}{AELV}{autonomous, electric, light-weight vehicle}
\newabbreviation{ETSI}{ETSI}{European Telecommunications Standards Institute}
\newabbreviation{EMC}{EMC}{Electromagnetic Compatibility}
\newabbreviation{EN}{EN}{European Standards}
\newabbreviation{F}{F}{Frequency}
\newabbreviation{FIT}{FIT}{Failure in Time} % der relative Anteil von fehlerhaften Elementen im Verhältnis zur Gesamtheit, also die relative Häufigkeit, mit der ein Fehler bei einem Produkt, einer Dienstleistung, einem Produktionsprozess oder der Arbeitsqualität auftaucht.
\newabbreviation{FMEA}{FMEA}{Failure Mode and Effects Analysis}
\newabbreviation{FuSa}{FuSa}{Functional Safety}
\newabbreviation{FSR}{FSR}{Functional Safety Requirement}
\newabbreviation{FTA}{FTA}{Fault Tree Analysis}
\newabbreviation{FZV}{FZV}{Vehicle Registration Law (German: Fahrzeug-Zulassungsverodnung)}
\newabbreviation{IEC}{IEC}{International Electrotechnical Commission}
\newabbreviation{ISO}{ISO}{International Organization for Standardization}
\newabbreviation{KBA}{KBA}{Federal Motor Transport Authority (German: Kraftfahrt-Bundesamt)}
\newabbreviation{OEMs}{OEMs}{Original Equipment Manufacturers}
\newabbreviation{QM}{QM}{Quality Management}
\newabbreviation{S}{S}{Severity}
\newabbreviation{StVG}{StVG}{Road Traffic Act (German: Straßenverkehrsgesetz)}
\newabbreviation{StVO}{StVO}{Road Traffic Regulations (German: Straßenverkehrs-Ordnung)}
\newabbreviation{StVZO}{StVZO}{Road Traffic Licensing Regulation (German: Straßenverkehrs-Zulassungs-Ordnung)}
\newabbreviation{TUEV}{TÜV}{Technical Monitoring Association (German: Technischer Überwachungsverein)}
\newabbreviation{UNECE}{UNECE}{United Nations Economic Commission for Europe}
\newabbreviation{VDA}{VDA}{Association of the Automotive Industry (German: Verband der Automobilindustrie)}
\newabbreviation{E/E/PE}{E/E/PE}{Electrical/Electronic/Programmable Electronic}
\newabbreviation{EU-R}{EU R}{European Union Regulation}
\newabbreviation{BPMN}{BPMN}{Business Process Model and Notation}
\newabbreviation{FBV}{FBV}{Vehicle Operation Regulation (German: Fahrzeug-Betriebs-Verordnung)}
\newabbreviation{EU}{EU}{European Union}
\newabbreviation{EC}{EC}{European Community}

% Customized part
\newabbreviation{ASR}{ASR}{Automatic Speech Recognition: an interdisciplinary subfield of computer science and computational linguistics that develops methodologies and technologies that enable the recognition and translation of spoken language into text by computers}
\newabbreviation{XLSR-53}{XLSR-53}{A large cross-lingual speech representation model pretrained in 53 languages}
\newabbreviation{Wav2vec 2.0}{wav2vec 2.0}{A framework for self-supervised learning of speech representations}
\newabbreviation{MT}{MT}{Machine Translation: a sub-field of computational linguistics that investigates the use of software to translate text or speech from one language to another}
\newabbreviation{masked_language_modeling}{Masked Language Modeling}{Giving a model a sentence and optimizing the weights inside the model to output the same sentence on the other side}
\newabbreviation{CNN}{CNN}{Convolutional Neural Network}
\newabbreviation{latent_speech_representations}{Latent Speech Representations}{Latent (or hidden) variables from empirical measurements, e.g. speech input}
\newabbreviation{Transformer}{Transformer}{A state-of-the-art architecture that aims to solve sequence-to-sequence tasks while handling long-range dependencies efficiently}
\newabbreviation{CE}{CE}{Cross-entropy}
\newabbreviation{E2E}{E2E}{End-to-End}
\newabbreviation{SOTA}{SOTA}{State-of-the-art}
\newabbreviation{GELU}{GELU}{Gaussian Error Linear Units: an activation function}
\newabbreviation{ReLU}{ReLU}{Rectified Linear Units: an activation function}
\newabbreviation{GMM}{GMM}{Gaussian Mixture Model}
\newabbreviation{HMM}{HMM}{Hidden Markov Model}
\newabbreviation{DNN}{DNN}{Deep Neural Network}
\newabbreviation{NN}{NN}{Neural Network}
\newabbreviation{MFCC}{MFCC}{Mel Frequency Cepstral Coefficients}
\newabbreviation{EM}{EM}{Expectation–maximization}
\newabbreviation{CART}{CART}{Classification and Regression Tree}
\newabbreviation{AM}{AM}{Acoustic model}
\newabbreviation{LM}{LM}{Language model}
\newabbreviation{YT}{YT}{YouTube}
\newabbreviation{fCE}{fCE}{Frame-wise Cross-entropy}
\newabbreviation{PPL}{PPL}{Perplexity}
\newabbreviation{OOV}{OOV}{Out-of-vocabulary}
\newabbreviation{BLSTM}{BLSTM}{Bidirectional Long-Short Term Memory}
\newabbreviation{LSTM}{LSTM}{Long-Short Term Memory}
\newabbreviation{ICE Loss}{ICE Loss}{Intermediate Cross-Entropy Loss}
\newabbreviation{IF Loss}{IF Loss}{Intermediate Focal Loss}
\newabbreviation{WER}{WER}{Word-error-rate}
\newabbreviation{WERR}{WERR}{Word-error-rate reduction}
\newabbreviation{RNN}{RNN}{Recurrent Neural Network}
\newabbreviation{Conformer}{Conformer}{Convolution-augmented Transformer for Speech Recognition}
\newabbreviation{GT}{GT}{Gammatone}
\newabbreviation{DCT}{DCT}{Discrete Cosine Transform}
\newabbreviation{DFT}{DFT}{Discrete Fourier Transform}
\newabbreviation{MLP}{MLP}{Multilayer Perceptron}
%\include{misc/glossary}

% Replace by useful values!
\makeTheTitlepage{
    organization={FH Aachen},
    title={Unsupervised Pre-Training for Vietnamese \\ Automatic Speech Recognition in the HYKIST Project},
    type={Bachelorarbeit},
    faculty={Fachbereich Medizintechnik und Technomathematik},
    department={Bachelorstudiengang Biomedizinische Technik},
    author={Le Duc Khai},
    matrikel={Matrikelnummer: 3089345},
    place=Jülich,
    date={Dezember 08, 2022},
    examiner={Prof. Dr. rer. nat. Ilya E. Digel, FH Aachen},
    secondExaminer={Christoph M. Lüscher M.Sc., RWTH Aachen},
    advisorFirst={Sen. Prof. Dr.-Ing. Hermann Ney, RWTH Aachen},
    advisorSecond={PD Dr. rer. nat. Ralf Schlüter, RWTH Aachen},
}

%\includepdf[pages={1}]{docs/ba_faber_punctuation}

\clearpage

\clearpage

\markboth{\textbf{Eigenständigkeitserklärung}}{\textbf{Eigenständigkeitserklärung}}
\mbox{}
%\vspace{0.5cm}\\

\textbf{Eigenständigkeitserklärung}

\noindent Ich erkläre hiermit, dass ich diese Bachelorarbeit selbstständig, ohne unzulässige Hilfe durch Dritte und ohne Benutzung anderer als der angegeben Hilfsmittel angefertigt habe. 
Insbesondere versichere ich, die aus anderen Quellen direkt oder indirekt übernommenen Daten und Konzepte sind unter Angabe der Quelle gekennzeichnet. 
Mir ist bekannt, dass meine Arbeit zum Zwecke eines Plagiatsabgleichs mittels einer Plagiatserkennungssoftware auf ungekennzeichnete Übernahme von fremdem geistigem Eigentum überprüft werden kann.

\vspace{1cm}
Jülich, \\
\vspace{1.0cm} \\
\rule{40mm}{0.2mm}\\
Le Duc Khai

\tableofcontents

\begin{spacing}{1.2}

    \chapter{Abstract}
\label{ch: Abstract}

In today's interconnected globe, moving abroad is more and more prevalent, whether it's for employment, refugee resettlement, or other causes.
Language difficulties between natives and immigrants present a common issue on a daily basis, especially in medical domain.
This can make it difficult for patients and doctors to communicate during anamnesis or in the emergency room, which compromises patient care. 
The goal of the HYKIST Project is to develop a speech translation system to support patient-doctor communication with \glsxtrshort{ASR} and \glsxtrshort{MT}.

\glsxtrshort{ASR} systems have recently displayed astounding performance on particular tasks for which enough quantities of training data are available, such as LibriSpeech \cite{panayotov2015librispeech}.
Building a good model is still difficult due to a variety of speaking styles, acoustic and recording settings, and a lack of in-domain training data.
In this thesis, we describe our efforts to construct \glsxtrshort{ASR} systems for a conversational telephone speech recognition task in the medical domain for Vietnamese language to assist emergency room contact between doctors and patients across linguistic barriers.
In order to enhance the system's performance, we investigate various training schedules and data combining strategies. We also examine how best to make use of the little data that is available. 
The use of publicly accessible models like \glsxtrshort{XLSR-53} \cite{xlsr53} is compared to the use of customized pre-trained models, and both supervised and unsupervised approaches are utilized using \glsxtrshort{Wav2vec 2.0} \cite{wav2vec2} as architecture.

\clearpage
    
    \chapter{Introduction}
\label{ch: Introduction}

\section{HYKIST Project}
\label{sec: HYKIST Project}

Migration to foreign countries is becoming more common in our globally connected world, whether for work, refugee movements, or other reasons.
As a result, language barriers between locals and foreigners are a common daily issue.
It is commonly known that, speaking with patients when they arrive at the hospital is crucial to their care. 
In medical care, a lack of or incorrect communication leads to underuse and misuse of medical services, lower quality of care, an increased rate of treatment errors, ineffective preventive measures for patients, and medical staff dissatisfaction.
The doctors then inquire about the patient's problems as well as his or her medical history. 
However, there are currently 20.8 million immigrants in Germany, with up to 30\% having only basic German language skills\footnote{\href{https://www.apptek.com/news/germanys-federal-ministry-of-health-awards-hykist-project-to-apptek-to-equip-critical-care-with-artificial-intelligence-driven-automatic-speech-translation-technology}{https://www.apptek.com/news/germanys-federal-ministry-of-health-awards-hykist-project-to-apptek-to-equip-critical-care-with-artificial-intelligence-driven-automatic-speech-translation-technology}}. 
If doctors and patients do not speak the same language, information communication is severely constrained, which has a negative impact on the patients' care. 
In the event that no common language is available, doctors can contact Triaphon which provides translators to aid communication between the patient and the doctor. 
These bi-lingual interpreters then assist in communication between the patient and the doctor.

In the HYKIST scenario, the doctor talks German to the patient, who speaks only Arabic or Vietnamese. 
Meanwhile, German and Arabic, or German and Vietnamese, are the languages spoken by the interpreters. 
The interpreters are not professional translators, instead, they are volunteers who contribute their time to the translation. 
This is problematic because the interpreters may require time to look up unfamiliar words, such as medical termini, or they may make a mistake.

The ultimate goal of the HYKIST project is to facilitate doctor-patient communication in a growing number of languages with the help of \glsxtrshort{ASR} and \glsxtrshort{MT} in order to meet the robust medical domain requirements via following steps: 
The interpreter is summoned via the hospital phone, which has an audio sampling rate of 8 kHz. 
We then create manual annotations with helps of our native-speaker volunteers. 
We investigate the use of additional outside-the-domain data for training as well as unsupervised methods because gathering project-specific data is an expensive and time-consuming operation.

\glsxtrshort{ASR} and \glsxtrshort{MT} technologies are linked with a dialogue system for initial anamnesis and integrated into an existing telecommunications platform for this purpose. 
First and foremost, the project collects dialogues in Arabic, Vietnamese, and German, which serve as the foundation for the development of algorithms and applications. 
During the project, the first technical tests for the accuracy and quality of the automated translations are already being performed. 
Following that, the overall system must be tested in a pilot test with clinical application partners for the area of emergency admissions and initial anamnesis in acute situations, as well as evaluated in a final clinical study for user acceptance.

The partners in the HYKIST Project are Triaphon\footnote{\href{https://triaphon.org/}{https://triaphon.org/}}, Fraunhofer Focus\footnote{\href{https://www.fokus.fraunhofer.de/en}{https://www.fokus.fraunhofer.de/en}} and AppTek GmbH\footnote{\href{https://www.apptek.com}{https://www.apptek.com}}.

\pagebreak

    \section{Motivation}
\label{sec: Motivation}

Large amounts of labeled training data benefit neural networks. 
However, labeled data is much more difficult to obtain in many settings than unlabeled data: current speech recognition systems require thousands of hours of transcribed speech to achieve acceptable performance, which is not available for the vast majority of the nearly 7,000 languages spoken globally \cite{ethnologue}. 
Learning solely from labeled examples is not comparable to human language acquisition: infants learn language by listening to adults around them - a process that necessitates the acquisition of good representations of speech.
Therefore, semi-supervised learning aims to work like the natural language acquisition of human.

Unsupervised and semi-supervised methods have been shown to be successful in \glsxtrshort{ASR} in recent years. 
\glsxtrshort{Wav2vec 2.0} \cite{wav2vec2}, in particular, has demonstrated excellent performance. 
\glsxtrshort{Wav2vec 2.0} is pre-trained using an unsupervised loss before being fine-tuned on labeled data.
The goal of the paper is to offer a framework for self-supervised learning of representations from raw audio data. 
This framework opens the door for speech recognition models to be used in a low-resource language like Vietnamese in medical domain where previously much more transcribed audio data was required to provide acceptable accuracy. 
The model is then fine-tuned on labeled data in a hybrid framework \cite{RASR-hybrid_vs_attention} after pre-training on unlabeled speech.

In the HYKIST Project, we want to utilize the \glsxtrshort{Wav2vec 2.0} model. 
One interesting aspect of \glsxtrshort{Wav2vec 2.0} is that the unsupervised pre-training is well suited for exploiting unlabeled multilingual data so that supervised training on a target language gains benefit from multilingual speech representations. 
In \cite{xlsr53}, the authors focused on learning representations from unlabeled data that generalize across languages in a multilingual scenario. 
They built on \glsxtrshort{Wav2vec 2.0} pretraining technique, in which a discrete vocabulary of \glsxtrshort{latent_speech_representations} is learned alongside contextualized speech representations. 
We can utilize their public model \glsxtrshort{XLSR-53} because it was unsupervised pretrained on 8 languages from Multilingual LibriSpeech \cite{multiling_librispeech}, 17 languages from the BABEL benchmark \cite{BABEL_dataset}, which is conversational telephone data with Vietnamese language included, as well as 36 languages from CommonVoice \cite{CommonVoice_dataset}, which is a corpus of read speech. 
With the exception of resource-rich languages, multilingual pretraining surpassed monolingual pretraining in most circumstances.

%\colorbox{Lavender}{Hybrid framework (Chris paper)}

%\colorbox{Lavender}{Aux loss}

\pagebreak
    \section{Related work}
\label{sec: Related_work}

Having been an established and effective method for \glsxtrshort{ASR}, hybrid modeling has made steady progress in recent years and outperformed \glsxtrfull{E2E} approach in most \glsxtrshort{ASR} situations \cite{RASR-hybrid_vs_attention}.
Besides, the recent introduction of novel neural encoders has been reported to significantly improve the performance \cite{facebook2020hybrid, google2020conformer, zeineldeen2022conformer}.
Other methods can also be used to achieve even greater improvements, like feature combination \cite{vieting2021waveform} or additional losses in the intermediate layers \cite{facebook2020dejavu}.
Furthermore, unsupervised approaches have grown in popularity due to their potential for high performance with little annotated data \cite{mohamed2022representation_review}. 
Semi-supervised learning was applied to an \glsxtrshort{ASR} task by \cite{deepmind2020cpc, facebook2019wav2vec, wav2vec2} by running unsupervised pre-training on a large unlabeled dataset, followed by fine-tuning on a small annotated dataset.
This technique can significantly reduce the amount of labeled data required to build \glsxtrshort{ASR} systems.
The successes sparked additional research into improving the modeling approach \cite{facebook2021hubert, facebook2022wav2vecaug} and analyzing which individual components contribute most to the performance \cite{livescu2021wav2vec_analysis}.
Besides, data used for pre-training and fine-tuning was deeply investigated as well, for example, in a domain-shift scenario \cite{robust_wav2vec2} in English language or using multilingual data for the sake of improvements on monolingual benchmarks \cite{xlsr53}.

Because the contrastive loss is computed solely on the input speech audio and does not require labels, it is especially simple to use for monolingual or multilingual data.
Therefore, a number of papers have begun to apply this loss for \glsxtrshort{ASR} research \cite{xlsr53, microsoft2021unispeech, zhang2021xlst, google2022just}.
Previously, supervised training with multilingual data could improve low resource languages by using a separate output layer for each language \cite{tuske2014multilingual}.
There has also been research specifically addressing medical domain tasks.
However, a common problem for medical \glsxtrshort{ASR} faced by researchers is difficult acoustic conditions and a lack of transcribed medical audio data \cite{edwards2017medicalspeech, chiu2018medconv, kar2021operation}.
Another difficulty likely to be met is the medical terminology.
In \cite{sakti2014towards}, a multilingual system for the medical domain is presented.
Another method for dealing with the medical domain is to correct \glsxtrshort{ASR} errors at the output level \cite{mani2020towardsmedical}.

To the best of our knowledge, unsupervised pretraining methods have mostly been investigated on well-known academic datasets, with no work done on applying them to difficult low-resource medical tasks. 
Furthermore, no previous work has been published that investigates the use of unsupervised pretraining methods for telephone speech directly on the 8kHz signal without resampling.
Besides, the analysis of different pretraining data combination and regularization for a medical \glsxtrshort{ASR} system has never been presented.

\pagebreak
    
    \chapter{Theory}
\label{ch: Theory}

\section{Hybrid ASR framework}

\subsection{Bayes theorem}

Given a sequence of acoustic observations $x_{1}^{T}$ whose length is $T$,  the most likely word sequence to be recognized is $w_{1}^{N}$.
A variety of subword units, such as phonemes, and the acoustic representation of the audio signal are connected through acoustic models.
In terms of probabilities, the relation $w^{*}$ between the acoustic and word sequence is described as: 

\begin{equation}
    w^{*} = \operatorname{arg}\max_{w_1^N} \, p(w_{1}^{N}|x_{1}^{T})
\end{equation}

As stated in the introduction, conventional \glsxtrshort{ASR} systems typically consist of a number of modules, including dictionaries, language models, and acoustic models.
By utilizing Bayes' Theorem to break out the posterior probability, it is possible to show the connections between them.
For the maximization, the probability $p(x)$ can be ignored because it just acts as a normalization and has no bearing on the outcome.

\begin{equation}
    p(w_{1}^{N}|x_{1}^{T}) = \frac{p(x_{1}^{T}|w_{1}^{N})p(w_{1}^{N})}{p(x_{1}^{T})} \propto p(x_{1}^{T}|w_{1}^{N})p(w_{1}^{N})
\end{equation}

\begin{equation}
w^{*} = \operatorname{arg}\max_{w_1^N}  \underbrace{p(x_{1}^{T}|w_{1}^{N})}_{\text{acoustic model}}\cdot\underbrace{p(w_{1}^{N})}_{\text{language model}}
\end{equation}

\subsection{Audio features}

The classification model uses features, which are representations taken from audio samples and used as input. 
There are many features, and they all show the spoken audio's frequency information.
Statistical models must learn some rather long-term dependencies within the input data due to the high resolution in the time-domain, which is often quite challenging and computationally expensive. 
As a result, we leverage acoustic features to simplify the signal while preserving the most crucial statistics.

\textbf{Mel-frequency cepstral coefficient (MFCC)}: The windowing of the signal, application of the \glsxtrfull{DFT}, calculation of the magnitude's log, warping of the frequencies on a Mel scale, and application of the inverse \glsxtrfull{DCT} are the main steps in the \glsxtrshort{MFCC} feature extraction technique. 
Below is a short explanation \cite{Rao_KE_2017} of each stage in the \glsxtrshort{MFCC} feature extraction process.

\begin{enumerate}
    \item Pre-emphasis: Filtering that highlights the higher frequencies is referred to as pre-emphasis. Its function is to balance the spectrum of spoken sounds, which roll off sharply at high frequencies.
    \item Frame blocking and windowing: Speech analysis over a short enough time span is required for stable acoustic features. The analysis must therefore always be performed on short segments where the speech signal is believed to be stationary.
    \item \glsxtrshort{DFT} spectrum: Each windowed frame is converted into magnitude spectrum by applying \glsxtrshort{DFT}
    \item Mel spectrum: The Fourier transformed signal is run through the Mel-filter bank, a collection of band-pass filters, to compute the Mel spectrum. A Mel is a unit of measurement based on the perceived frequency by human ears.
    \item \glsxtrfull{DCT}: Because the vocal tract is smooth, there is a tendency for adjacent bands' energy levels to correlate. When the converted Mel frequency coefficients are applied to the \glsxtrshort{DCT}, a set of cepstral coefficients are generated.
    \item Dynamic MFCC features: Since the cepstral coefficients only include data from a single frame, they are frequently referred to as static features. By computing the first and second derivatives of the cepstral coefficients, additional information on the temporal dynamics of the signal is gained.
\end{enumerate}

\textbf{Gammatone features}: The Gammatone filter \cite{Aertsen_Olders_Johannesma_1981}, which is intended to mimic the human auditory filter, is the foundation for Gammatone features. 
They were initially presented for large vocabulary \glsxtrshort{ASR} in \cite{schlueter:icassp07}. 
A filterbank of Gammatone filters with center frequencies sampled from the Greenwood function \cite{Greenwood_1990} is applied after pre-emphasizing the speech signal.
Below is a summary of each stage in the Gammatone feature extraction process:

\begin{enumerate}
    \item Typically, a Hanning window of 25 ms width with 10 ms shifts is used to perform the temporal integration of the absolute values of the filter outputs.
    \item A spectral integration with a 9-channel window and a 4-channel shift followed.
    \item (10th root or log) compression was performed, followed by cepstral decorrelation resulting in 16 cepstral coefficients. 
    \item Following the use of the 10th root compression, a discrete cosine transform (DCT)-based cepstral decorrelation and normalizing methods are used.
\end{enumerate}

\textbf{Extracted features from raw waveform}: The features from raw waveform encoder are extracted by \glsxtrshort{CNN} feature encoder. 
First, the feature encoder's raw waveform input is normalized to zero mean and unit variance.
The feature encoder contains seven blocks and the temporal convolutions in each block have 512 channels with strides (5,2,2,2,2,2,2) and kernel widths (10,3,3,3,3,2,2). 
Besides, layer normalization \cite{layer_normalization}, and the \glsxtrshort{GELU} activation function \cite{gelu} are also applied.
This results in an encoder output frequency of 49 hz with a stride of about 20ms between each sample, and a receptive field of 400 input samples or 25ms of audio. 
The convolutional layer modeling relative positional embeddings has kernel size 128 and 16 groups.

\subsection{Acoustic modeling}

When modeling the probability $p(x_1^T|w_1^N)$, the length of time sequence $T$ and of word sequence $N$ are often not the same because $N$ is usually much smaller than $T$.
The alignment between the acoustic observations $x_1^T$ and labels $w_1^N$ is unknown and commonly even unclear.
The \glsxtrfull{HMM} is a statistical model that introduces a latent alignment by states $s_1^T$ and subsequently modeling the probability of $x_1^T$ for a given alignment to $w_1^N$ \cite{Baum1967AnIW}.
The probability $p(x_1^T|w_1^N)$ is then calculated by adding all possible alignments between the acoustic observation and the labels.
Assuming conditional independence of observations when states are given and that states only depend on their predecessor, this sum results in the equation below:

\begin{align}
	\label{eq:hmm}
	p(x_1^T|w_1^N) = \sum_{[s_1^T]}\prod_{t=1}^Tp(x_t, s_t|s_{t-1}, w_1^N) =
	\sum_{[s_1^T]}\prod_{t=1}^T\underbrace{p(s_t|s_{t-1}, w_1^N)}_{\text{transition prob.}}\cdot
	\underbrace{p(x_t|s_t, s_{t-1}, w_1^N)}_{\text{emission prob.}}
\end{align}

A widely accepted simplification is to make the assumption that the last state for the emission probability is independent such that:

\begin{align}
	\label{eq:simplify_hmm}
	p(x_t|s_t, s_{t-1}, w_1^N) = p(x_t|s_t, w_1^N)
\end{align}

The transition model calculates the probabilities of moving from one state to the next. 
The emission probability models the probability of an acoustic observation based on the current and previous states. 
When the probability is simplified, it only depends on the current state.
The transition model can have several topologies, but the \textit{0-1-2} topology is the most commonly used.
The topology is state-independent and has different transition probabilities: staying in the current state, jumping to the next state, or jumping to the second next state.
By jumping faster or slower in time, the jump and stay property allows the alignment of labels and acoustic observations to adjust.
The emission model calculates the probability of an acoustic observation in the current and previous states.

\textbf{Context-Dependent Phone}: Because a language's vocabulary is typically very large, modeling words directly in the classification is impractical.
Phonemes, on the other hand, are frequently used for subword modeling.
For better learning, the acoustic articulation of a phoneme is determined by its surroundings, for example  the beginning, the middle and the ending part.
As a result, multiple phonemes are combined to create triphone or allophone labels.

\textbf{\glsxtrfull{CART}} \cite{Beulen98automaticquestion}: However, because of the cubic number of phonemes, these are a large class of labels. 
The possible triphones are greater than the number of observed triphones.
Therefore, some share the same \glsxtrshort{GMM} model.
\glsxtrshort{CART} is a decision tree used to cluster triphones that can share the same \glsxtrshort{GMM} model.
To reduce the number of labels, allophones are clustered using a \glsxtrshort{CART}, and the subsequent clusters are used as labels.

\textbf{Baum–Welch algorithm}: In practical training of \glsxtrshort{HMM}, inferring the parameters of the \glsxtrshort{HMM} is not simple and cannot be done manually.
An automated data-driven approach based on the \glsxtrfull{EM} algorithm is used instead, with a dataset of acoustic observations with transcriptions.
Because the best alignment between acoustic observations and transcriptions is not always available, the \glsxtrshort{EM} algorithm is initially leveraged with a sub-optimal linear alignment.
The observation model and alignment are then iteratively optimized using the steps below:

\begin{enumerate}
	\item Maximization: Estimate the model parameters using the previously obtained alignment by maximizing the log-likelihood function.
	\item Expectation: Using the parameters from step 1, estimate a new alignment.  
	\item Get back to step 1 until the model fully converges.
\end{enumerate}

\textbf{\glsxtrshort{GMM}/\glsxtrshort{HMM}}: The \glsxtrshort{HMM} can be used to model the transition between phones and the corresponding observable.
A widely used approach is modelling the emission probabilities for each label with a parametrized \glsxtrshort{GMM}, resulting the \glsxtrshort{GMM}/\glsxtrshort{HMM} method.
The \glsxtrshort{GMM} is a weighted sum over $K$ normal distributions

\begin{align}
	p(x_t|s_t, s_{t-1}, w_1^N) = \sum_{i=1}^K c_i \cdot \mathcal{N}(x_t|\mu_i, \sigma_i^2),
\end{align}

resulting in a multimodal emission probability with parameters $\mu_{i}, \sigma_{i}$ and mixture weights $c_i$ for $i\in\llbracket1,K\rrbracket$. 
The mixture weights are non-negative and sum up to unity.
Using the simplification in Equation \ref{eq:simplify_hmm} the state $s_{t-1}$ can be additionally dropped. 

\textbf{\glsxtrshort{DNN}/\glsxtrshort{HMM}}: Another approach that has been popular is modelling the posterior probability $p(a_{s_t}|x_1^T)$ discriminatively.
Usually \glsxtrfull{DNN} is leveraged for this purpose, resulting in the \glsxtrshort{DNN}/\glsxtrshort{HMM} approach.
The purpose of \glsxtrshort{GMM}/\glsxtrshort{HMM} system is to generate alignments for the training of \glsxtrshort{DNN}/\glsxtrshort{HMM} system \cite{RASR-hybrid_vs_attention}.
The emission probability in the \glsxtrshort{HMM} can afterwards be calculated by applying Bayes rule such that:

\begin{align}
	p(x_1^T|a_{s_t}) = \frac{p(a_{s_t}|x_1^T)p(x_1^T)}{p(a_{s_t})}.
\end{align}

The probability $p(a_{s_t})$ can be estimated as the relative frequency of $a_{s_t}$.
In order to simplify the Bayes decision rule, the probability $p(x_1^T)$ is constant and therefore can be removed.

\subsection{Language modeling}

In a hybrid system, we use the 4-gram count based \glsxtrfull{LM}, using Kneser-Ney Smoothing algorithm \cite{kneser_ney_lm}. 
The \glsxtrshort{LM}s employed all use full-words in the first-pass decoding \cite{beck2019lstm}. 
In other words, lattice rescoring is not performed in the second-pass decoding.

In order to deal with multiple monolingual text corpora, the first step is to create an \glsxtrshort{LM} for each monolingual text corpus. 
Following that, we use a weighting process to combine the \glsxtrshort{LM}s into a single \glsxtrshort{LM}, yielding one \glsxtrshort{LM} for Vietnamese language.

\subsection{Decoding}

In order to recognize the speech given the acoustic observations, the \glsxtrshort{AM} and \glsxtrshort{LM} need to be combined following the Bayes decision rule, resulting in:

\begin{align}
	w_1^N = \operatorname{arg}\max_{N,w_1^N}p\Bigl(\prod_{n=1}^Np(w_n|w_{n-m}^{n-1}) \cdot
	\sum_{[s_1^T]}\prod_{t=1}^Tp(x_t,s_t|
	s_{t-1}, w_1^N)\Bigr)
\end{align}

With dynamic programming, this maximization can be solved by Viterbi algorithm which recursively computes the maximum path in $O(k^{2}T)$ where $k$ and $T$ are vocabulary size and sequence length respectively.
The Viterbi approximation can be applied as

\begin{align}
	w_1^N = \operatorname{arg}\max_{N,w_1^N}p\Bigl(\prod_{n=1}^Np(w_n|w_{n-m}^{n-1}) \cdot
	\max_{[s_1^T]}\prod_{t=1}^Tp(x_t,s_t|s_{t-1}, w_1^N)\Bigr),
\end{align}
so that the optimization reduces to a best-path problem in the alignment graph of all possible predicted words to the acoustic observations.
Besides, beam search (\glsxtrshort{AM} and \glsxtrshort{LM} pruning) is used in the searching process which only focuses on the most promising predicted words at each time step \cite{ortmanns1997word}.

\subsection{Recognition Performance}

The \glsxtrfull{WER} is a widely used indicator of how well an \glsxtrshort{ASR} system is performing. 
The percentage of words that were incorrectly predicted is shown by this number. 
The \glsxtrshort{ASR} system performs better with a lower value; a \glsxtrshort{WER} of 0 equals a perfect result. \glsxtrshort{WER} can be calculated as:

\begin{equation}
    \text{WER} = \frac{\text{Substitutions} + \text{Insertions} + \text{Deletions}}{\text{Reference words}}
\end{equation}
    \section{Neural network}

A neural network is a set of algorithms that attempts to recognize underlying relationships in a set of data using a process that mimics how the human brain works.
Neural network contains layers of interconnected nodes. 
Each node is known as a perceptron.

\subsection{Multilayer perceptron}

By adding one or more hidden layers, we can get around the drawbacks of linear models.
Stacking a lot of fully connected layers on top of one another is the simplest approach to accomplish this. 
Up until we produce outputs, each layer feeds into the layer above it. The first layers serve as our representation, and the top layer serves as our linear predictor. This design is frequently referred to as a \glsxtrfull{MLP}.

\begin{figure}[!h]
    \centering
    \includegraphics[width=0.7\textwidth]{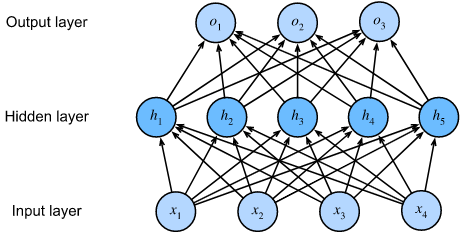}
    \caption{An MLP with a hidden layer of 5 hidden units \cite{zhang2021dive}}
    \label{multilayer_perceptron}
\end{figure}

This \glsxtrshort{MLP} has 4 inputs, 3 outputs, and 5 hidden units in its hidden layer.
Because the input layer does not require any computations, producing outputs with this network necessitates implementing computations for both the hidden and output layers; thus, the number of layers in this \glsxtrshort{MLP} is 2. 
It should be noted that both layers are fully connected. 
Every input influences every neuron in the hidden layer, and every neuron in the output layer influences every neuron in the hidden layer.

We denote by the matrix $X \in R^{n \times d}$ a minibatch of $n$ examples where each example has $d$ inputs (features). 
For a one-hidden-layer \glsxtrshort{MLP} whose hidden layer has $h$ hidden units, we denote by $H \in R^{n \times h}$ the outputs of the hidden layer, which are hidden representations. 
Since the hidden and output layers are both fully connected, we have hidden-layer weights $W^{(1)} \in R^{d \times h}$ and biases $b^{(1)} \in R^{1 \times h}$ and output-layer weights $W^{(2)} \in R^{h \times q}$ and biases $b^{(2)} \in R^{1 \times q}$. 
This allows us to calculate the outputs  of the one-hidden-layer MLP as follows:

\begin{equation}
\begin{aligned}
   H = X W^{(1)} + b^{(1)} \\
   O = H W^{(2)} + b^{(2)}
\end{aligned}
\end{equation}

To fully realize the potential of multilayer architectures, one more key component is required: a nonlinear activation function to be applied to each hidden unit after the affine transformation.
For instance, a popular choice is the ReLU (Rectified Linear Unit) activation function \cite{relu} $\sigma (x) = \max (0, x)$ operating on its arguments element-wise. 
The outputs of activation functions are called activations. 
In general, with activation functions in place, our \glsxtrshort{MLP} cannot be collapsed into a linear model.

\begin{equation}
\begin{aligned}
   H = \sigma (X W^{(1)} + b^{(1)}) \\
   O = H W^{(2)} + b^{(2)}
\end{aligned}
\end{equation}

\subsection{Training a neural network}

\textbf{Epoch}: one iteration where the model sees the whole training set to update its weights.

\textbf{Mini-batch gradient descent}: during the training phase, updating weights is usually not based on the whole training set at once due to computation complexities or one data point due to noise issues. 
Instead, the update step is done on mini-batches, where the number of data points in a batch is a hyperparameter (batch size) that we can tune.

\textbf{Loss function}: In order to quantify how a given model performs, the loss function $L$ is usually used to evaluate to what extent the actual outputs $y$ are correctly predicted by the model outputs $z$.

\textbf{Cross-entropy loss}: In the context of binary classification in neural networks, the cross-entropy loss $L(z,y)$ is commonly used and is defined as follows:
\begin{equation}
    L(z,y) = -[y \log (z) + (1-y) \log (1-z)]
\end{equation}

\textbf{Forward propagation}: The calculation and storage of intermediate variables (including outputs) for a neural network from the input layer to the output layer is referred to as forward propagation (or forward pass).

\textbf{Backpropagation}: The method of calculating the gradient of neural network parameters is known as backpropagation. In short, the method traverses the network in reverse order, from the output to the input layer, using calculus' chain rule. While calculating the gradient with respect to some parameters, the algorithm stores any intermediate variables (partial derivatives).

\textbf{Updating weights}: In a neural network, weights are updated as follows:

\begin{itemize}
    \item[] Step 1: Take a batch of training data and perform forward propagation (feedforward) to compute the loss.
    \item[] Step 2: Backpropagate the loss to get the gradient of the loss with respect to each weight
    \item[] Step 3: Use the gradients to update the weights of the network.
\end{itemize}

\begin{figure}[!h]
    \centering
    \includegraphics[width=1.0\textwidth]{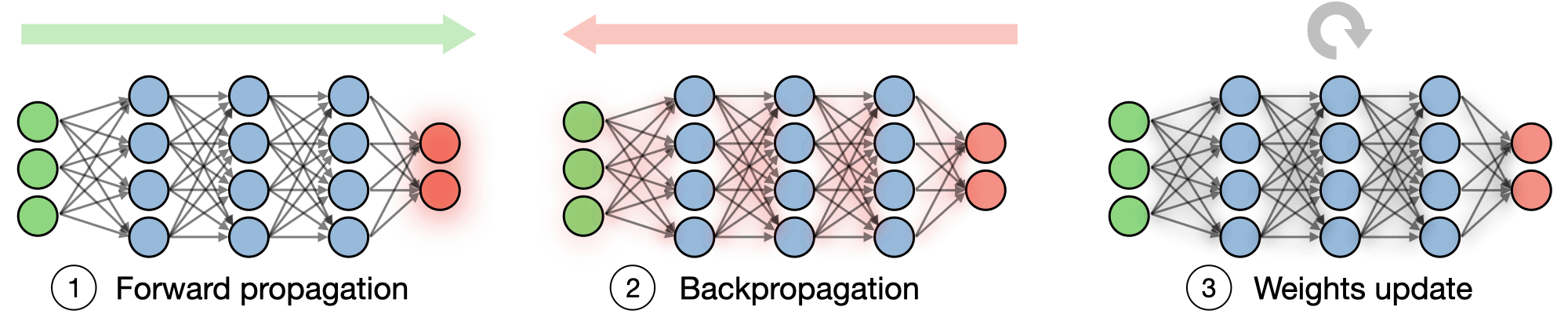}
    \caption{Updating weights in a neural network \cite{amidi2018deep}}
    \label{backpropagation}
\end{figure}

\subsection{Parameter tuning}

\begin{figure}[!h]
    \centering
    \includegraphics[width=1.0\textwidth]{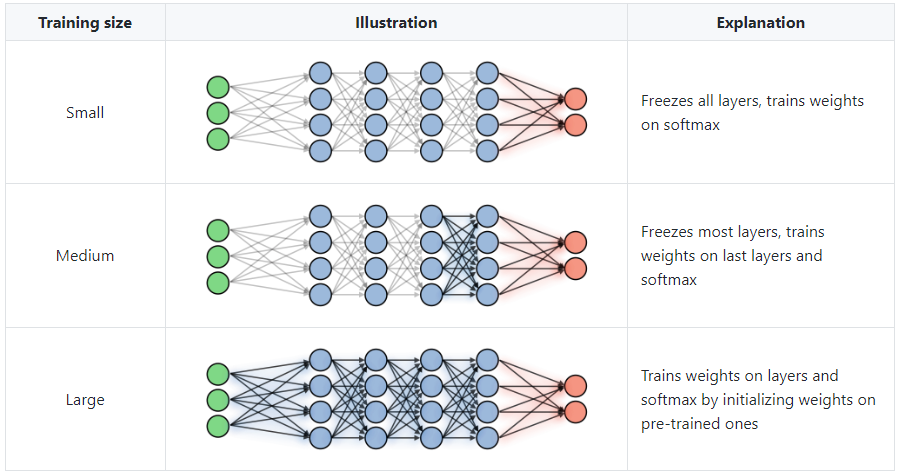}
    \caption{Transfer learning strategy \cite{amidi2018deep}}
    \label{transferlearning}
\end{figure}

Weights initialization:
\begin{itemize}
    \item[] Xavier initialization \cite{xavier_init}: Rather than simply randomizing the weights, Xavier initialization allows for initial weights that take into account characteristics that are unique to the architecture. 
    Weights and inputs are centered at zero, while biases are initialized as zeros.
    \item[] Transfer learning: It is frequently useful to leverage pre-trained weights from massive datasets that took days/weeks to train and apply them to our use case. Figure \ref{transferlearning} shows some options for leveraging data, depending on how much we have:
\end{itemize}

\begin{figure}[!h]
    \centering
    \includegraphics[width=0.95\textwidth]{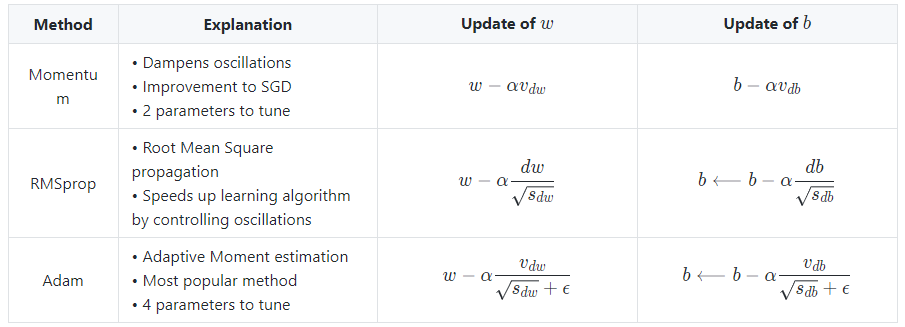}
    \caption{Adaptive learning rates methods \cite{amidi2018deep}}
    \label{adaptiveLR}
\end{figure}

Optimizing convergence:
\begin{itemize}
    \item[] Learning rate: indicates how quickly the weights are updated. 
    It can be fixed or changed adaptively. 
    The most popular method at the moment is Adam \cite{kingma2015adam}, which is a method that adapts the learning rate.
    \item[] Adaptive learning rates: Allowing the learning rate to vary when training a model can help to reduce training time while also improving the numerical optimal solution. While the Adam optimizer is the most commonly used technique, the following in figure \ref{adaptiveLR} are also useful:
\end{itemize}

Regularization:

\begin{itemize}
    \item[] Dropout \cite{dropout}: to avoid overfitting the training data by removing neurons with probability $p > 0$. 
    It forces the model to avoid relying too heavily on specific sets of features.
    \item[] Weight regularization: Regularization techniques are typically used on the model weights to ensure that the weights are not too large and that the model is not overfitting the training set.
    \item[] Early stopping: to halt training as soon as the validation loss reaches a plateau or begins to rise.
    \item[] SpecAugment \cite{park2019specaugment}: Rather than augmenting the input audio waveform, SpecAugment applies an augmentation policy directly to the audio spectrogram (i.e., an image representation of the waveform).
    The spectrogram is altered by warping it in time, masking blocks of consecutive frequency channels, and masking blocks of utterances in time.
    These augmentations are chosen to help the network to be robust against deformations in the time direction, partial loss of frequency information and partial loss of small segments of speech of the input.
\end{itemize}

\subsection{Convolutional Neural Network}

Architecture of a traditional \glsxtrfull{CNN} is generally composed of the following layers:
\begin{itemize}
    \item[] Convolution layer (CONV): This layer employs filters that perform convolution operations while scanning the input $I$ in terms of its dimensions. 
    The filter size $F$ and stride $S$ are two of its hyperparameters. 
    The resulting output $O$ is referred to as a feature map or an activation map.
    \item[] Pooling layer (POOL): a downsampling operation used after a convolution layer to achieve spatial invariance. 
    Max and average pooling, in particular, are types of pooling that take the maximum and average value, respectively.
    \item[] Fully connected layer (FC): works with a flattened input, with each input connected to all neurons. 
    FC layers, when present, are typically found near the end of \glsxtrshort{CNN} architectures and can be used to optimize objectives such as class scores.
\end{itemize}

\subsection{Recurrent Neural Network}

\begin{figure}[!h]
    \centering
    \includegraphics[width=1.0\textwidth]{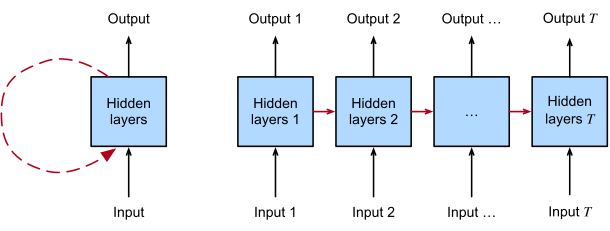}
    \caption{Recurrent connections are depicted on the left as cyclic edges. The RNN is unfolded over time steps on the right. Recurrent edges are computed synchronously, while conventional connections span adjacent time steps. \cite{zhang2021dive}}
    \label{rnn}
\end{figure}

\glsxtrfull{RNN} is a deep learning model that captures the dynamics of sequences through recurrent connections, which can be viewed as node cycles in a network (connections between nodes can create a cycle).
\glsxtrshort{RNN}s are unrolled across time steps (or sequence steps) using the same underlying parameters at each step. 
While standard connections are used synchronously to propagate activations from one layer to the next at the same time step, recurrent connections are dynamic, passing information across adjacent time steps.
As illustrated in Figure \ref{rnn}, \glsxtrshort{RNN}s are feedforward neural networks in which the parameters of each layer (both conventional and recurrent) are shared across time steps.

\subsection{Bidirectional Long Short-Term Memory}

The most popular designs include mechanisms to mitigate \glsxtrshort{RNN}s' infamous numerical instability, as exemplified by vanishing and exploding gradients.
We present the key concepts underlying the most successful \glsxtrshort{RNN} architectures for sequence, which are based on two papers published in 1997.

\glsxtrfull{LSTM} \cite{lstm1997} is the first paper to introduce the memory cell, a unit of computation that replaces traditional nodes in a network's hidden layer.
With these memory cells, networks can overcome training difficulties encountered by previous recurrent networks. 
To avoid the vanishing gradient problem, the memory cell keeps values in each memory cell's internal state cascading along a recurrent edge with weight 1 across many successive time steps. 
A set of multiplicative gates assists the network in determining which inputs to allow into the memory state and when the memory state's content should influence the model's output.
Given memory cell $c_t$, input gate $i_t$, forget gate $f_t$, output gate $o_t$ associated with weight matrices $W_j$, $U_j$ and weight vector $b_j$ where $j \in \{i, f, o, c\}$, \glsxtrshort{LSTM} is described as:

\begin{equation}
\begin{aligned}
   i_{t} &= \operatorname{sigmoid}_{g}(W_{i}x_{t} + U_{i}h_{t-1} + b_{i}) \\
   f_{t} &= \operatorname{sigmoid}_{g}(W_{f}x_{t} + U_{f}h_{t-1} + b_{f}) \\
   o_{t} &= \operatorname{sigmoid}_{g}(W_{o}x_{t} + U_{o}h_{t-1} + b_{o}) \\
   c_{t} &= f_{t} \odot c_{t-1} + i_{t} \odot \operatorname{sigmoid}_{c}(W_{c}x_{t} + U_{c}h_{t-1} + b_{c}) \\
   h_{t} &= o_{t} \odot \operatorname{sigmoid}_{h}(c_{t}) 
\end{aligned}
\end{equation}

The second paper, Bidirectional \glsxtrfull{RNN} \cite{brnn1997}, describes an architecture that uses information from both the future (subsequent time steps) and the past (preceding time steps) to determine the output at any point in the sequence. 
This is in contrast to previous networks, in which only previous input could influence output. 
Bidirectional \glsxtrshort{RNN}s have become a mainstay in audio sequence labeling tasks, among many others. 
Fortunately, the two innovations are not mutually exclusive and have been successfully combined for phoneme classification and handwriting recognition.

\subsection{Transformer}

\begin{figure}[!h]
    \centering
    \includegraphics[width=0.5\textwidth]{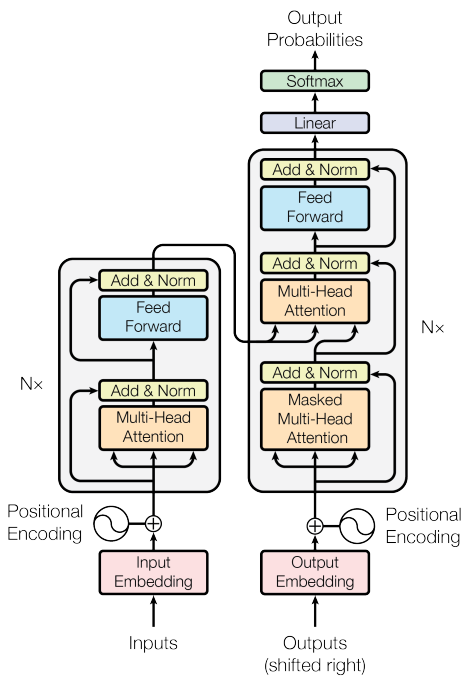}
    \caption{The Transformer model architecture \cite{Transformer}}
    \label{transformer_blocks}
\end{figure}

The Transformer employs the encoder-decoder architecture, as shown in the left and right halves of Figure \ref{transformer_blocks}, with stacked self-attention and point-wise, fully connected layers for both the encoder and decoder.

The encoder is built up from N identical layers. 
Each layer is divided into two sub-layers. 
The first is a multi-head self-attention mechanism, and the second is a simple, fully connected feed-forward network that is positionally connected. 
Following layer normalization \cite{layer_normalization}, a residual connection \cite{DeepResidualLearning} is used around each of the two sub-layers.

\textbf{Attention}: A query and a set of key-value pairs are mapped to an output by an attention function, where the query, keys, values, and output are all vectors. 
The output is computed as a weighted sum of the values, with the weight assigned to each value determined by the query's compatibility function with the corresponding key.

\textbf{Scaled Dot-Product Attention}: The input consists of
queries and keys of dimension $d_{k}$, and values of dimension $d_{v}$.
The query's dot products are computed with all keys, divided by $\sqrt d_{k}$, and a softmax function is applied to get the weights on the values.
In practice, we compute the attention function on a set of queries at the same time, which we pack into a matrix $Q$. 
The keys and values are also packed into matrices $K$ and $V$. 
We compute the output matrix as follows:

\begin{equation}
    \operatorname{Attention}(Q, K, V ) = \operatorname{softmax}(\frac{QK^{T}}{\sqrt d_{k}}) V
\end{equation}

\textbf{Multi-Head Attention}: Instead of performing a single attention function with $d_{model}$-dimensional keys, values and queries, we perform the attention function in parallel on each of the projected versions of queries, keys, and values, yielding $d_{v}$-dimensional output values.
These are concatenated and projected again, yielding the final values:

\begin{equation}
    \operatorname{MultiHead}(Q, K, V ) = \operatorname{Concat}(\mathrm{head}_{1}, ..., \mathrm{head}_{h})W^{O}
\end{equation}

where: $\mathrm{head}_{i} = \operatorname{Attention}(QW_{i}^{Q}, KW_{i}^{K}, VW_{i}^{V})$ 

and the projections are parameter matrices $W_{i}^{Q} \in R^{d_{\mathrm{model}} \times d_{k}}$, $W_{i}^{K} \in R^{d_{\mathrm{model}} \times d_{k}}$, $W_{i}^{V} \in R^{d_{\mathrm{model}} \times d_{v}}$ and $W^{O} \in R^{hd_{v} \times d_{\mathrm{model}}}$,

$h$ is the number of attention heads.

    \section{Semi-supervised learning}

Semi-supervised learning is a method of machine learning in which a small amount of labeled data is combined with a large amount of unlabeled data during training. 
Semi-supervised learning is intermediate between unsupervised (no labeled training data) and supervised learning (with only labeled training data). 
It is an example of weak supervision.

When combined with a small amount of labeled data, unlabeled data can significantly improve learning accuracy. 
Acquiring labeled data for a learning problem frequently necessitates the use of a skilled human agent (e.g., to transcribe an audio segment in \glsxtrshort{ASR} tasks).
The cost of labeling may thus make large, fully labeled training sets unfeasible, whereas acquiring unlabeled data is relatively inexpensive. Semi-supervised learning can be extremely useful in such situations.

\subsection{Wav2vec 2.0}

Due to self-supervised training, \glsxtrshort{Wav2vec 2.0} is one of the current \glsxtrshort{SOTA} models for \glsxtrshort{ASR}. 
This is a relatively novel concept in this sector. We can pre-train a model on unlabeled data, which is always more accessible, using this method of training. 
The model can then be fine-tuned for a specific purpose using a specific dataset.

\begin{figure}[hbtp]
    \centering
    \includegraphics[width=0.8\textwidth]{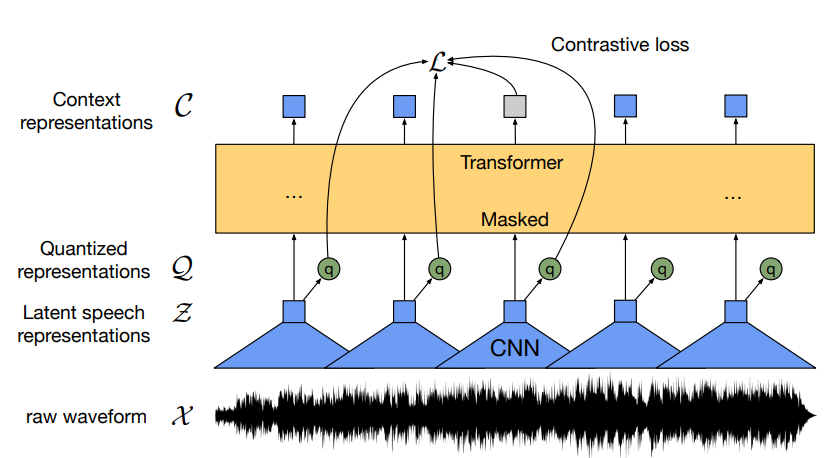}
    \caption{Illustration of our framework which jointly learns contextualized speech representations and an inventory of discretized speech units.}
    \label{speech_representation_wav2vec2}
\end{figure}

The model consists of a multi-layer convolutional feature encoder $f: X \rightarrow Z$ that receives raw audio $X$ as input and produces \glsxtrshort{latent_speech_representations} ${z}_1,..., {z}_T$ for $T$ time steps. 
They are then supplied into a \glsxtrshort{Transformer} $g: Z \rightarrow C$, which generates representations ${c}_1,..., {c}_T$ that capture data from the full sequence. 
In the self-supervised objective, the output of the feature encoder is discretized to $q_t$ using a quantization module $Z \rightarrow Q$ to represent the objectives (Figure \ref{speech_representation_wav2vec2}). 
The approach constructs context representations over continuous speech representations, and self-attention captures dependencies throughout the whole sequence of latent representations.

\textbf{Feature encoder}: The encoder is made up of many blocks that include temporal convolution, layer normalization \cite{layer_normalization}, and the \glsxtrshort{GELU} activation function \cite{gelu}. 
The encoder's raw waveform input is normalized to zero mean and unit variance. 
The number of time-steps T that are input to the \glsxtrshort{Transformer} is determined by the encoder's total stride.

\textbf{Contextualized representations with Transformers}: The feature encoder's output is sent into a context network that uses the \glsxtrshort{Transformer} architecture \cite{Transformer}. 
We utilize a convolutional layer that acts as a relative positional embedding instead of fixed positional embeddings that encode absolute positional information. 
We implement layer normalization after adding the convolution output followed by a \glsxtrshort{GELU} to the inputs.

\textbf{Contrastive learning}: Contrastive learning is a notion that involves the input being altered in two ways. 
The model is then trained to recognize whether two input transformations are still the same item. 
The \glsxtrshort{Transformer} layers are the first method of transformation in \glsxtrshort{Wav2vec 2.0}; the second is quantization. 
In more technical terms, we would like to get such a context representation $c_t$ for a masked latent representation $z_t$ in order to guess the proper quantized representation $q_t$ among alternative quantized representations.

\begin{figure}[hbtp]
    \centering
    \includegraphics[width=0.8\textwidth]{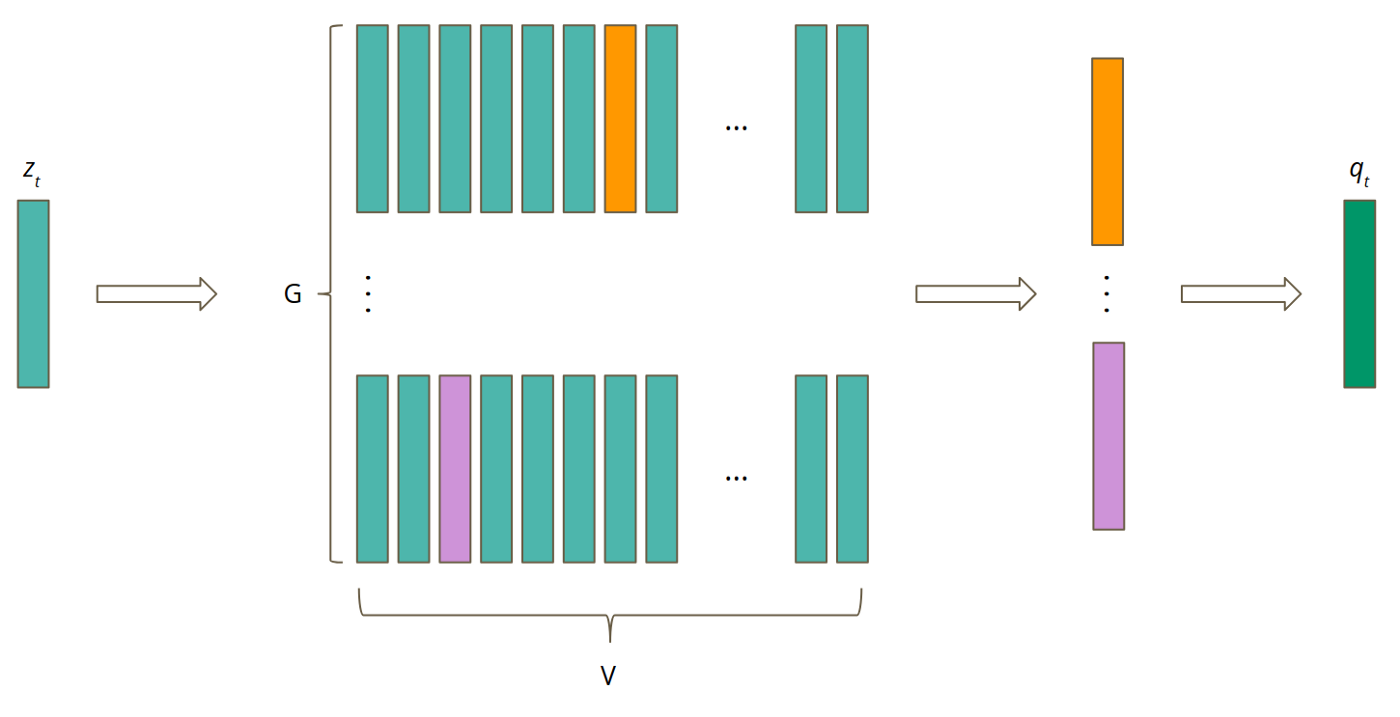}
    \caption{Quantization process: For each codebook, the best entry is extracted and concatenated with each other (from orange to purple entry)}
    \label{quantization_process}
\end{figure}

\textbf{Quantization module}: Quantization is a process of converting values from a continuous space into a finite set of values in a discrete space \cite{wav2vec2_towardsdatascience}. 
A language's number of phonemes is limited. 
Furthermore, the number of posible phoneme pairs is limited. 
It means that the same \glsxtrshort{latent_speech_representations} can correctly represent both of them. 
Furthermore, because the quantity is limited, we can design a codebook that contains all potential phoneme combinations. 
The quantization process then involves selecting the appropriate code word from the codebook. 
However,the total number of conceivable sounds is enormous. 
To make it easier to learn and use, we use product quantization \cite{product_quantization} to discretize the output of the feature encoder $z$ to a finite set of speech representations for self-supervised training. 
This choice yielded positive results, which acquired discrete units first and then contextualized representations. 
Concatenating quantized representations from several codebooks is what product quantization is all about. 
We take one item from each codebook and concatenate the resulting vectors $e_1,...,e_G$ (Figure \ref{quantization_process}), then perform a linear transformation $R^d \rightarrow R^f$ to get $q \in R^f$, given $G$ codebooks or groups with $V$ entries $e \in R^{V \times d/G}$.

\subsection{Cross-lingual speech representation}

Cross-lingual learning seeks to create models that use data from other languages to improve performance.
By pretraining \glsxtrshort{Transformer} blocks with multilingual masked language models, unsupervised cross-lingual representation learning has shown great success \cite{lample2019cross, BERT}. 
The authors in \cite{xlsr53} studied cross-lingual speech representations by extending \glsxtrshort{Wav2vec 2.0} \cite{wav2vec2} to the cross-lingual setting. 
Their method teaches a single set of quantized latent speech representations that are shared by all languages.
They pre-trained \glsxtrshort{XLSR-53} on 56k hours of speech data from 53 languages (including Vietnamese language), then evaluated it on 5 languages from the BABEL benchmark (conversational telephone data) \cite{BABEL_dataset} and 10 languages from CommonVoice \cite{CommonVoice_dataset} - a corpus of read speech.

\subsection{In-domain Match Level and Diversity Level}

In this part, to better and easier analyze the effect of pre-training data on the performance of cross-lingual and domain-shift experiments, we introduce 2 new concepts, namely \textbf{"In-domain Match Level"} and \textbf{"Diversity Level"}. 

\textbf{In-domain Match Level}: Given 3 datasets A, B and C, where A is the target telephone dataset used for recognition, B is also recorded by the telephone but its conversation is different from A's and C is the audio book recordings. 
The dataset B is more overlapped with the A than the C because both A and B are telephone recordings, so the \textbf{In-domain Match Level} of B is higher than the one of C. 
In general, the \textbf{In-domain Match Level} is determined by the similarity between recording conditions, naturalness and conversational topics.

\textbf{"Diversity Level"}: Given another dataset D, which is recorded by more speakers with more diverse accents than B and C, then the \textbf{Diversity Level} of D is the highest compared to the rest. 
To some extent, the \textbf{Diversity Level} of the multilingual dataset is higher than the monolingual one because the first is able to represent more learnable phonemes which are likely to be helpful to target language in semi-supervised learning.

    \chapter{Experiments}
\label{ch: Experiments}

\section{Data}

The first difficulty faced during the research in the HYKIST project is the lack of medical telephone speech dataset.
Having a small medical dataset - HYKIST, we therefore use HYKIST only for the recognition and use in-house non-medical telephone speech dataset for training. 
This poses a challenge to reach a high-performance ASR because of the mismatch in training and recognition datasets.
In addition, real-life dataset like HYKIST is difficult to be accurately transcribed by ASR models because of background noises, variation of speaking speed, unfamiliar pronunciation of medical terms...

\subsection{HYKIST data}

Our HYKIST project partner Triaphon recorded conversations between three people: a patient, a doctor, and an interpreter. 
The patient communicates in the non-German language - Arabic or Vietnamese - while the doctor communicates in German. 
The interpreter is fluent in both languages and assists the patient and doctor in communicating. 
In HYKIST, we have unique accents, foreign-born accents, from both interpreter and patient sides.
This directly makes HYKIST more difficult for machines and humans to transcribe, leading understandable bad recognition performance.
We received the audio recordings and had our transcribers perform speech transcription within the recordings. 
We divide the audio data into two sets: dev and test, with no speaker overlap between the two.

The data statistics for the dev and test sets for each individual language can be seen in Table \ref{table:data_stats}.
We only have a limited amount of data because we create it ourselves.
Furthermore, the number of speakers is limited, resulting in a low level of diversity in the testing data.
This may result in over-optimization of the evaluation data.
To address the impact of the data issues, we obtained additional training data from our industry partner Apptek and other sources.

\begin{table}[!h]
\centering
\begin{adjustbox}{width=\columnwidth,center}
\begin{tabular}{|l|c|c|c|c|c|c|c|} 
\hline
Language                    & Dataset                 & Usage                                                     & \# Spks & Hours  & Domain                                                                       & \begin{tabular}[c]{@{}c@{}}In-domain \\match\end{tabular} & \begin{tabular}[c]{@{}c@{}}Diversity \\level\end{tabular}  \\ 
\hline
Arabic                      & In-house                & pretr.                                                    & 3379    & 786    & Tel., Conv.                                                                  & Medium                                                    & Medium                                                     \\ 
\hline
German                      & In-house                & pretr.                                                    & 1723    & 177    & Tel., Conv.                                                                  & Medium                                                    & Medium                                                     \\ 
\hline
\multirow{5}{*}{Vietnamese} & In-house                & \begin{tabular}[c]{@{}c@{}}pretr., \\finetu.\end{tabular} & 2240    & 219    & Tel., Conv.                                                                  & Medium                                                    & Medium                                                     \\ 
\cline{2-8}
                            & \multirow{3}{*}{HYKIST} & adapt                                                     & 1       & 1      & \multirow{3}{*}{\begin{tabular}[c]{@{}c@{}}Tel., Conv.,\\Med.~\end{tabular}} & \multirow{3}{*}{High}                                     & \multirow{4}{*}{Low}                                       \\ 
\cline{3-5}
                            &                         & dev                                                       & 3       & 3      &                                                                              &                                                           &                                                            \\ 
\cline{3-5}
                            &                         & test                                                      & 2       & 2      &                                                                              &                                                           &                                                            \\ 
\cline{2-7}
                            & YouTube                 & pretr.                                                    & -       & 1.204  & Read books                                                                   & Low                                                       &                                                            \\ 
\hline
\multirow{2}{*}{Multi}      & In-house*               & pretr.                                                    & 7342    & 1.182  & Tel., Conv.                                                                  & Medium                                                    & \multirow{2}{*}{High}                                      \\ 
\cline{2-7}
                            & XLSR-53                 & pretr.                                                    & -       & 56.000 & Various                                                                      & Low                                                       &                                                            \\
\hline
\end{tabular}
\end{adjustbox}
\caption{Data statistics for acoustic data. *The multilingual in-house training dataset is the combination of the Arabic, German and Vietnamese ones listed above. Domain: Telephone (Tel.), Conversational (Conv.), Medical (Med.).}
\label{table:data_stats}
\end{table}

\subsection{In-house data}

AppTek, an industry partner, supplied us with annotated 8kHz conversational telephone speech data.
The audio data was collected during telephone conversations between customers and various call centers.
Table \ref{table:data_stats} displays the data statistics for the training sets for each of the three languages.
We can see that the amount of training data available varies between languages.

We also have speakers with accents and/or dialects for the Arabic and Vietnamese data.
For the Arabic data, we have four different datasets with distinct dialects: Syrian, Lebanese, Gulf, and Egyptian.
Besides, our Vietnamese dataset has dominantly 2 accents, Northern and Central Vietnamese, and a very small fraction of Southern Vietnamese accent.
The speakers with accents in the Vietnamese data are combined into a single dataset.

\subsection{YouTube}

We collected Vietnamese audio data from \glsxtrfull{YT} under Fair Use Policies\footnote{\href{https://support.google.com/youtube/answer/9783148}{https://support.google.com/youtube/answer/9783148}} in addition to our annotated datasets.
The domain in question is purely read speech, such as podcasts, audiobooks, radio stories, or something similar.
Pre-processing was done manually by removing non-speech parts such as music and noise, leaving only speech.
The audio files were then divided into 10-30 second segments.
Table \ref{table:data_stats} displays the data statistics for the web scraped data.
During data collection, we headed to the balance of accents and genders.
Therefore, the dataset is divided into Northern and Southern accents, yielding four subsets: Northern Female (518h), Northern Male (213h), Southern Female (290h) and Southern Male (183h).

%\subsection{VLSP}

%The VLSP dataset is a spontaneous-reading speech dataset provided by Association for Vietnamese Language and Speech Processing.
%The 2021 version\footnote{\href{https://vlsp.org.vn/vlsp2021/eval/asr}{https://vlsp.org.vn/vlsp2021/eval/asr}} includes 280 hours of transcribed data in the general domain and 400 hours of untranscribed data.
%The dataset is recorded in different real scenarios e.g., meeting conversation, lecture speech. 
%To evaluate our models, we use \cite{Duy_Khanh_Finetune_Wav2vec_2_0_2022}'s model which was trained on 100h of VLSP dataset, together with CommonVoice and VIVOS described below.

\subsection{CommonVoice Vietnamese}

We obtain the Vietnamese dataset from the massively-multilingual speech corpus  \cite{ardila2020commonvoice}. 
We use the data version 9.0\footnote{\href{https://commonvoice.mozilla.org/en/datasets}{https://commonvoice.mozilla.org/en/datasets}}, which includes 17 hours of noisy read speech data recorded by the large number of volunteer speakers.
The dataset is split into train/dev/test set. 
We evaluate our models by directly recognizing on dev and test sets.

\subsection{VIVOS}

VIVOS \cite{vivos_dataset} is a clean Vietnamese read speech corpus consisting of 15 hour recordings.
We obtain the dataset\footnote{\href{https://ailab.hcmus.edu.vn/vivos}{https://ailab.hcmus.edu.vn/vivos}} split into train/test sets.
We evaluate our models by directly recognizing on test set.
The test set includes 19 speakers and 48 minutes of duration in total.

\subsection{Monolingual text data}

Apptek, our project partner, provided monolingual text data for all three languages.
Text from various sources is included in the data.
The number of running words for each language is shown in Table \ref{table:LM_stats}.

\subsection{Domain}

As shown in Table \ref{table:data_stats} the data spans several domains.
The HYKIST project's target domain is medical conversational telephone speech.
The training data does not cover this specific domain.
This domain mismatch in our data is highlighted.
By listening to the audios and comparing them to our target domain, we can determine the in-domain match and diversity level.

    \section{Lexicon and language model}

% Please add the following required packages to your document preamble:
% \usepackage{multirow}
\begin{table}[!ht]
\setlength{\tabcolsep}{0.9em}
\centering
\begin{tabular}{|c|c|c|c|c|c|} 
\hline
\# words & vocab & \multicolumn{2}{c|}{dev} & \multicolumn{2}{c|}{test}  \\ 
\hline
in train & size  & OOV   & PPL              & OOV   & PPL                \\ 
\hline
500M     & 11k   & 0.1\% & 67               & 0.2\% & 69                 \\
\hline
\end{tabular}
\caption{ 4-gram LM for Vietnamese.}
\label{table:LM_stats}
\end{table}

\subsection{Lexicon}

The Babel project\footnote{\href{https://www.iarpa.gov/research-programs/babel}{https://www.iarpa.gov/research-programs/babel}} provided us the initial lexicon for the Vietnamese language.
The training lexicon is then created by extending the initial lexica with the toolkit Sequitur Grapheme-To-Phoneme\footnote{\href{https://github.com/sequitur-g2p/sequitur-g2p}{https://github.com/sequitur-g2p/sequitur-g2p}} \cite{bisani2008g2p}.
We supplement the lexicon with medical terms provided by our project partner Triaphon in order to decode the HYKIST data.
The final recognition lexica for Vietnamese are 11k in size as shown in Table \ref{table:LM_stats}.

\subsection{Language model}

\glspl{LM} used are 4-grams and use entire words. 
We create our \glspl{LM} using the training pipeline from the SRILM toolkit \cite{stolcke2002srilm}.
The first step is to create a \glsxtrshort{LM} for each monolingual text corpus separately. 
Then, using a weighting procedure, we merge all \glspl{LM}  into a single \glsxtrshort{LM}, producing one \glsxtrshort{LM} for Vietnamese language.
Using the development text, interpolation weights can be determined by giving highest weight to the source language models that have the lowest perplexity on the specified development set.

Table \ref{table:LM_stats} demonstrates how the \glsxtrshort{LM} performs. 
Vietnamese \glsxtrshort{LM} achieves a \glsxtrfull{PPL} of 67 and a \glsxtrfull{OOV} rate of 0.1\% on dev set.

    \section{Acoustic model}
\label{ch: Acoustic_model}

In this part, our experimental setups for acoustic models are described.
We use the toolkit 
RETURNN\footnote{\href{https://github.com/rwth-i6/returnn}{https://github.com/rwth-i6/returnn}} \cite{doetsch2016returnn}
for supervised training experiments and
Fairseq\footnote{\href{https://github.com/facebookresearch/fairseq}{https://github.com/facebookresearch/fairseq}} \cite{facebook2019fairseq}
for unsupervised \glsxtrshort{Wav2vec 2.0} training.
The recognition is done by using
RASR\footnote{\href{https://github.com/rwth-i6/rasr}{https://github.com/rwth-i6/rasr}} \cite{rybach2011rasr}.
We convert the Fairseq models to RETURNN models with an automatic
conversion toolkit\footnote{\href{https://github.com/rwth-i6/pytorch-to-returnn-converter}{https://github.com/rwth-i6/pytorch-to-returnn-converter}}.
We will release all training and decoding configurations
online\footnote{\href{https://github.com/rwth-i6/returnn-experiments}{https://github.com/rwth-i6/returnn-experiments}}\footnote{\href{https://github.com/rwth-i6/i6-experiments}{https://github.com/rwth-i6/i6-experiments}}.

\subsection{Supervised-only models}

The training schedule for Vietnamese language's basic systems
are similar and simply differ in the specifics. Assuming all models, we generate alignments obtained through the use of a \glsxtrshort{GMM}/\glsxtrshort{HMM} procedure will be utilized as labels for neural network training.
In a supervised setting using \glsxtrshort{fCE}, all models are trained from scratch. The labels used in the \glsxtrshort{AM} modeling are context-dependent phonemes, more specific triphones.
With 4501 \glsxtrshort{CART} labels in the end, we use a \glsxtrshort{CART} to tie the states. 
We employ the 40-dimensional Gammatone features as the \glsxtrshort{AM}'s input \cite{schlueter:icassp07}. 

There is no pre-training, so the fine-tuning begins with a random initialization.
All the fine-tunings from scratch takes 33 epochs.
We use two distinct neural \glsxtrshort{AM} architectures: \glsxtrshort{Transformer} \cite{Transformer}, and \glsxtrfull{BLSTM} \cite{hochreiter1997long}. 

\textbf{BLSTM}: We strictly adhere to the training recipe in \cite{RASR-hybrid_vs_attention} for the \glsxtrshort{BLSTM} model. 
The \glsxtrshort{BLSTM} uses 5 layers and 512 per-direction units. 
The following hyperparameters are used for fine-tuning: 
The initial learning rate is set at $0.0005$, followed by a hold phase, and finally an exponential decay with decay factor of 0.8 in order to control the learning rate based on CE development set scores.
In addition, we use Adam optimizer with Nesterov momentum (Nadam) \cite{dozat2016incorporating}.
Furthermore, a dropout of 10\% is applied to all modules and batch shuffling is turned off.
A batch size of 40000 frames is employed.
The SpecAugment \cite{park2019specaugment} algorithm is used for entire model training with masking of
50\% in the time dimension and 10\% in the feature dimension. 
This leads to the \glsxtrshort{BLSTM} size of 25M parameters.

\textbf{Transformer}: Our \glsxtrshort{Transformer} training schedule was obtained from \cite{zeineldeen2022conformer, zeineldeen2022robustconformer}. 
\glsxtrshort{Transformer} has 12 blocks.
The attention dimension of each Multi-Head Self-Attention module is 768 with 12 attention heads. 
The dimension of the feed-forward module is 1536 with \glsxtrshort{ReLU} working as an activation function.
The following hyperparameters are used for fine-tuning: 
The initial learning rate is set at $10^{-5}$ and a linear warm-up phase to $10^{-4}$ is used, followed by a hold phase, and finally an exponential decay with decay factor of 0.9 until the minimum learning of $10^{-7}$ is reached.
In addition, we use Adam optimizer with Nesterov momentum (Nadam) \cite{dozat2016incorporating}.
Furthermore, a dropout of 10\% is applied to all layers of encoder network and we use batch size of 8000 frames.
Batches are constructed with shuffled data.
The SpecAugment \cite{park2019specaugment} algorithm is used for entire model training with masking of
50\% in the time dimension and 10\% in the feature dimension. 
This leads to the \glsxtrshort{Transformer} size of 90M parameters.

\subsection{Models using unsupervised pre-training}

\textbf{XLSR-53}: We look into using a publically accessible model, \glsxtrshort{XLSR-53} \cite{xlsr53}, in addition to pre-training our own models on our specific data.
We utilize the checkpoint that was not fine-tuned to any language\footnote{\href{https://github.com/facebookresearch/fairseq/tree/main/examples/wav2vec}{https://github.com/facebookresearch/fairseq/tree/main/examples/wav2vec}}. 
This was pre-trained on 56k hours of speech data from 53 different languages for 19 epochs.
Additionally, we explore with initializing the \glsxtrshort{Wav2vec 2.0} pre-training on our custom data using the \glsxtrshort{XLSR-53} model, followed by corresponding fine-tuning.
Note that 16kHz data were used to train the \glsxtrshort{XLSR-53}. 
We shorten the stride of one \glsxtrshort{CNN} layer in the feature extractor to half because we work with 8kHz telephone conversation.
In this method, we receive features at the desired frame rate while reducing the down-sampling factor from the waveform to the feature frames by a factor of 2.

\textbf{Pretraining cases}: For each pretrained model we divide into the following cases. 
\begin{enumerate}
	\item Instead of a custom pre-training with our available datasets, \glsxtrshort{XLSR-53} is applied directly for the fine-tuning.
	\item Pre-training on our available datasets from scratch.
	\item The parameters are initialized with the 
	\glsxtrshort{XLSR-53} checkpoint and the pre-training is done with our available datasets.
	We call this type of pre-training continued pretraining.
\end{enumerate}

\textbf{Wav2vec 2.0 architectures}: We use the topologies from \glsxtrshort{Wav2vec 2.0} for the experiments with unsupervised pre-training \cite{wav2vec2} and customize our own topologies into:
\textit{Base}, \textit{Large} and \textit{Large}\textsubscript{1-8}.
All architectures work with the raw audio waveform and have a feature extractor that uses 7 \glsxtrshort{CNN} layers. 
However, the \textit{Large} architecture has an encoder stack made up of 24 \glsxtrshort{Transformer} layers with dimension of the feed-forward module being 1024 and the number of attention heads being 16. 
\textit{Base} only has 12 \glsxtrshort{Transformer} layers with dimension of the feed-forward module being 768 and the number of attention heads being 12.
\glsxtrshort{Wav2vec 2.0} Large model trained on multilingual data makes up the \glsxtrshort{XLSR-53} model \cite{xlsr53}. 
The 24 \glsxtrshort{Transformer} layers employed in the Large architecture place a heavy burden on the GPU's memory. 
Training times are dramatically increased when GPU memory is traded for a smaller batch size. 
We suggest discontinuing the \glsxtrshort{Wav2vec 2.0} Large network after the 
8\textsuperscript{th} \glsxtrshort{Transformer} block and referring to the model as
\textit{Large}\textsubscript{1-8}\xspace 
in order to mitigate.
We discovered that an optimal trade-off between a large enough batch size and a model size that still fits into memory is 8 layers.
The cut-off reduces the model size of the full
architecture \textit{Large} from 317M parameters to 115M of \textit{Large}\textsubscript{1-8}\xspace and is
therefore much closer to 95M parameters of \textit{Base} architecture.
In addition to the difference between architectures, 

\textbf{Pretraining}: During pretraining we employ the proposed hyperparameters in the \glsxtrshort{XLSR-53} paper \cite{xlsr53} but apply the learning rate of \glsxtrshort{Wav2vec 2.0} for the monolingual pre-trainings.
The pre-trainings are done for 300 epochs if there is nothing mentioned otherwise.
A linear warm-up is used during the first 30 epochs until the learning rate reaches 0.0005 and then a linear decay starts.
The mini-batch size in the existing Fairseq implementation is specified in samples of the waveform.
For both \textit{Base} and \textit{Large}\textsubscript{1-8} we use a dropout of 10\% in the feature extractor, 
5\% in the encoder and 10\% in the latent representations between the feature extractor and encoder.
We do not apply dropout to pre-trainings with the \textit{Large} architecture.
a \glsxtrshort{NN} is pre-trained on unlabeled data using the contrastive loss and diversity loss as described in \cite{wav2vec2} using the \glsxtrshort{Wav2vec 2.0} framework.

\textbf{Finetuning}: To finetune the acoustic model, we use the training system described in \cite{RASR-hybrid_vs_attention} to create a baseline \glsxtrshort{GMM}/\glsxtrshort{HMM} model for Vietnamese language. 
This model is used to generate alignments of the speech data with the \glsxtrshort{CART} labels for the \glsxtrshort{DNN} system. 
The hybrid model's \glsxtrshort{NN} is trained on these alignments in a supervised manner using the \glsxtrfull{fCE} loss. 
An application of a two-stage training configuration is made when using unsupervised pre-training.
After pretraining, the \glsxtrshort{NN} is then fine-tuned by adding a softmax output layer, initializing with a checkpoint from pre-training, training with the \glsxtrshort{fCE} loss on labeled data, and using the same alignment as in the fully supervised scenario.
The following hyperparameters are used for fine-tuning:
The initial learning rate is set to $10^{-5}$ and uses a linear warm-up phase to $10^{-4}$ followed by a hold phase and afterwards ends with exponential decay of 0.9.
The \glsxtrshort{Wav2vec 2.0} SpecAugment variant introduced in \cite{wav2vec2} is used with the masking done by choosing independent random starting points in the time/feature dimension and the subsequent 10/64 steps are masked.
We employ a mini-batch size of 1875 frames with length of 10ms, leading to the audio of 18.75 seconds.
Furthermore, a gradient noise of 10\% is used and we also apply a dropout of 5\% to all layers of both feature extractor and Transformer encoder network.

In addition to dropout \cite{dropout}, we also investigate the performance of other regularization techniques like the intermediate loss, L2 and On-off Regularization described in \ref{subsec: Intermediate_loss}, \ref{subsec: L2_regularization} and \ref{subsec: On_off_Regularization}

\subsection{Data augmentation}

In this thesis, apart from the use of SpecAugment \cite{park2019specaugment} stated above, we also use other data augmentation techniques in the pretraining stage.

The augmentation was exclusively done using the speed perturbation \{90\%, 110\%, 115\%\} \cite{speed_perturbation}, random pitch perturbation \{-350:-250; 250:350\} \cite{pitch_perturbation} and reverberation perturbation \cite{reverb_perturbation}. 
We did not go further to analyze if these augmentation options are the most optimal.

\subsection{Intermediate loss}
\label{subsec: Intermediate_loss}

Our intermediate loss setups are based on \cite{facebook2020dejavu, zeineldeen2022conformer}. 
Besides, we have 2 variants of intermediate loss, namely \glsxtrfull{ICE Loss}, which uses \glsxtrfull{CE} loss and \glsxtrfull{IF Loss}, which replaces\glsxtrfull{CE} loss with focal loss \cite{focal_loss}.

\textbf{\glsxtrshort{ICE Loss}}: We conducted multiple experiments with intermediate loss scales ranging in \{0.1, 0.2, 0.3, 0.4, 0.5\} and dropout \cite{dropout} values ranging in \{0.05, 0.1\}. 
We saw that the combination of loss scale 0.3 and dropout value 0.1 yielded the best results for all pretrained models and architectures, so we take this as default for all next experiments.

\textbf{\glsxtrshort{IF Loss}}: We experimented with 3 ways of integrating focal loss into the vanilla intermediate loss setup: only in the network \glsxtrshort{CE} output layer, only in the intermediate loss layer and in both of them. 
We found that putting the focal loss in both 2 positions yielded better result.
To find a good focal loss value, we conducted experiments with multiple values in \{1.5, 2.0, 2.5, 3.0\}. 
The higher the value is, the more on labels the network is forced to "focus". 
We saw that the 2 values \{1.5, 2.0\} did not make difference in results, while for higher focal values \{2.5, 3.0\}, the model gained more benefits on in-house training set but hurt the performance on out-domain recognition test sets. 
We highly recommend the use of focal value 2.0 so that the model generalizes on all different test sets.

\subsection{L2 regularization}
\label{subsec: L2_regularization}

To find good values of L2 regularization \cite{L2_regularization}, we used grid-search technique. 
We tested the value ranging in \{0.01, 0.005, 0.001, 0.0005, 0.0001\} to see the resulting \glsxtrfull{WER}.
Each pretraining model and architecture has its own unique L2 value to work best.
We put L2 regularization at all linear layers in the network.

\subsection{On-off Regularization}
\label{subsec: On_off_Regularization}

To further improve the accuracy performance of \glsxtrshort{IF Loss}, we introduce a new regularization technique called "On-off Regularization technique". 
We turn off all regularizations (Dropout, SpecAugment and \glsxtrshort{IF Loss}) in the first stage of training (3-10 first epochs).
We call this stage "Off Regularization".
We then reset the learning rate and turn all regularizations back on in the second stage of training, which we call "On Regularization".
The second stage of training ends when the model is fully converged.
    
    \chapter{Experimental results}
\label{ch: Experimental_results}

\section{Supervised baselines}

% Please add the following required packages to your document preamble:
% \usepackage{multirow}
\begin{table}[!ht]
\centering
\begin{adjustbox}{width=0.6\columnwidth,center}
\begin{tabular}{|l|c|c|} 
\hline
\multicolumn{1}{|c|}{\multirow{2}{*}{AM}} & \multicolumn{2}{c|}{WER [\%]}  \\ 
\cline{2-3}
\multicolumn{1}{|c|}{}                    & Hykist dev & Hykist test       \\ 
\hline
GMM                                       & 62.2       & 59.7              \\ 
\hline
BLSTM                                     & 32.9       & 38.4              \\ 
\hline
Transformer                               & 31.0       & 35.1              \\
\hline
\end{tabular}
\end{adjustbox}
\caption{\Glspl{WER} [\%] for supervised-only baselines on Vietnamese HYKIST data \cite{luescher2022:hykist}. Models are trained on the monolingual in-house data. The labels are context-dependent phonemes (\glsxtrshort{CART} state tying) with their size being 4501.}
\label{table:supervised_baselines}
\end{table}

The baseline for Vietnamese is trained using the relevant in-house 8kHz monolingual telephone speech data.
The performance of the baseline \glsxtrshort{ASR} systems is displayed in Table \ref{table:supervised_baselines}.
The intrinsic difficulty of the language and the data causes the systems to function differently.
We believe there are various causes for this.
Due to the natural flow of speakers, the Vietnamese transcriptions are hard to reach high quality.
Additionally, Vietnamese also incorporates accented speech which is even difficult for native speakers to fully understand. 
Furthermore, the accent mismatch between Vietnamese fine-tuning and recognition data is also a major factor to the degradation of performance.
Our Vietnamese in-house dataset has dominantly 2 native accents, Northern and Central Vietnamese, and a very small fraction of Southern Vietnamese native accent, while HYKIST, because of being a simulation dataset, has unique accents - foreign-born accents - from both interpreter and patient sides.

On the HYKIST data, switching from a \glsxtrshort{GMM}/\glsxtrshort{HMM} framework to a hybrid \glsxtrshort{HMM} framework with a \glsxtrshort{RNN}-\glsxtrshort{BLSTM} results in a reduction of \glsxtrshort{WER} from 62.2\% and 59.7\% to 32.9\% and 38.4\% on dev and test set respectively.
Besides, the \glsxtrshort{WER}s continue decreasing to 31.0\% and 35.1\% by replacing \glsxtrshort{BLSTM} with  \glsxtrshort{Transformer} encoder.
    \section{Unsupervised Pre-training}
\label{sec: unsupervised_pretraining}

\subsection{Monolingual pre-training}

Table \ref{table:monoling_pretraining} shows the outcomes from models pretrained on monolingual data.
The number of pre-training epochs is decided upon using the best downstream \glsxtrshort{WER} on Vietnamese.

\begin{table}[!ht]
\centering
\begin{tabular}{|c|c|c|c|c|} 
\hline
\multicolumn{2}{|c|}{Pre-training}                                                           & Fine-tuning         & \multicolumn{2}{c|}{WER [\%]}  \\ 
\hline
Data (hours)                                                          & Epochs               & Epochs              & Hykist dev & Hykist test       \\ 
\hline
None                                                                  & None                 & 33                  & 32.1       & 36.6              \\ 
\hline
\begin{tabular}[c]{@{}c@{}}Viet. in-house \\~(219h)\end{tabular}      & 100                  & \multirow{4}{*}{26} & 31.4       & 33.4              \\ 
\cline{1-2}\cline{4-5}
\begin{tabular}[c]{@{}c@{}}Aug. Viet. in-house \\(1168h)\end{tabular} & \multirow{3}{*}{300} &                     & 31.0       & 32.3              \\ 
\cline{1-1}\cline{4-5}
\begin{tabular}[c]{@{}c@{}}Viet. YT \\(1168h)\end{tabular}            &                      &                     & 29.8       & 35.2              \\ 
\cline{1-1}\cline{4-5}
\begin{tabular}[c]{@{}c@{}}Viet. in-house + YT \\(1168h)\end{tabular} &                      &                     & 25.3       & 27.2              \\
\hline
\end{tabular}
\caption{\glspl{WER} {[}\%{]} for models pretrained on monolingual data. All fine-tunings use the \textit{Large}\textsubscript{1-8} architecture and are trained until full convergence on Vietnamese in-house data and the recognition is done on HYKIST. All pre-trainings have been done with random initialization. Pre-training data "None" in the 3rd row means fine-tuning from scratch with \glsxtrshort{Wav2vec 2.0} \textit{Large}\textsubscript{1-8} architecture.}
\label{table:monoling_pretraining}
\end{table}

Even though no additional data is included for pre-training here, pre-training on the monolingual in-house data for Vietnamese reveals a reduction of \glsxtrshort{WER}s from 32.1\% and 36.6\% to 31.4\% and 33.4\% on dev and test set respectively.
This proves that on \glsxtrshort{Wav2vec 2.0} architecture, the unsupervised pretraining helps the \glsxtrshort{WER} performance.

Next, when we pretrain with the augmented in-house data, we achieve a small improvement to 31.0\% and 32.3\% on dev and test set respectively.
This shows that data augmentation for pretraining is helpful.

We then examine the impact of pre-training on the \glsxtrfull{YT} data for Vietnamese, which results improvements to 29.8\% and 35.2\%.
Although \glsxtrshort{YT} data is much more than the in-house data (1168h compared to 219h), both results seem to similar in terms of the average result on dev and test set. 
This proves that having more data is not always helpful, because of 2 reasons.
The first reason is that the domain of the in-house data is closer to that of HYKIST (both of them are telephone domain), while the domain mismatch between \glsxtrshort{YT} and HYKIST is larger (read speech compared to telephone speech).
Another reason is that \glsxtrshort{YT} data has less speakers, leading to worse generalization while pretraining.

The greatest significant improvement is achieved by combining the in-house and \glsxtrshort{YT} data leading to a reduction of \glsxtrshort{WER}s to 25.3\% and 27.2\%.
This is the best result produced using solely monolingual data. 
Because we substitute 200 hours of \glsxtrshort{YT} with in-house data, the amount of pre-training data used here is comparable to that of only \glsxtrshort{YT} pre-training. 
This result proves that a diversity of domains and speakers in the pretraining stage is necessary for better performance on test sets.

\subsection{Multilingual pre-training}

\begin{table}[!ht]
\centering
\begin{tabular}{|c|c|c|c|c|} 
\hline
\multicolumn{3}{|c|}{Pre-training}                                                                                       & \multicolumn{2}{c|}{WER [\%]}  \\ 
\hline
Init                    & Data (hours)                                                            & Epochs               & Hykist dev & Hykist test       \\ 
\hline
\multirow{2}{*}{random} & \begin{tabular}[c]{@{}c@{}}Viet. in-house + YT\\(1168h)\end{tabular}    & \multirow{2}{*}{300} & 25.3       & 27.2              \\ 
\cline{2-2}\cline{4-5}
                        & \begin{tabular}[c]{@{}c@{}}Multilingual in-house \\(1168h)\end{tabular} &                      & 26.8       & 28.7              \\ 
\hline
\textit{XLSR-53}\textsubscript{1-8}     & None                                                                    & None                 & 27.6       & 31.9              \\
\hline
\end{tabular}
\caption{\glspl{WER} {[}\%{]} for models using unsupervised pretraining on multilingual data compared to pretraining on monolingual data. All fine-tunings use the \textit{Large}\textsubscript{1-8} architecture and are trained until full convergence on Vietnamese in-house data and the recognition is done on HYKIST. The 2nd model is pretrained on our multilingual in-house dataset, and the 3rd uses \textit{XLSR-53}\textsubscript{1-8} to directly finetune on Vietnamese in-house data.}
\label{table:multiling_pretraining}
\end{table}

We then examine models that have already been multilingually pre-trained in Table \ref{table:multiling_pretraining}. 
For Vietnamese dev/test, combining the Arabic, German, and Vietnamese in-house data to create a custom multilingual pre-training significantly outperforms the non-pretraining baseline, at \glsxtrshort{WER}s of 26.8\% and 28.7\% on dev and test set respectively. 
However, the monolingual combination of in-house and \glsxtrshort{YT} data is still better for Vietnamese, at 25.3\% and 27.2\% on dev and test set respectively.
These results reject \cite{xlsr53}'s conclusion where multilingual pretraining is proved to outperform monolingual pretraining.

Strong increases of \glsxtrshort{WER}s can also be seen by fine-tuning only utilizing the \textit{XLSR-53}\textsubscript{1-8} checkpoint. 
With the exception of the Vietnamese test set, where it is up to 11\% worse, it performs only relatively worse than the custom pre-training on the multilingual in-house data, at \glsxtrshort{WER}s of 27.6\% and 31.9\%.
This may be due to the absence of 8kHz data in the pre-training of \textit{XLSR-53}. 
Nevertheless, adopting it in a fast and simple manner can result in considerable benefits.

\subsection{\glsxtrshort{XLSR-53} as pre-training initialization}

\begin{table}[!ht]
\centering
\begin{tabular}{|c|c|c|c|c|} 
\hline
\multicolumn{3}{|c|}{Pre-training}                                                                                                           & \multicolumn{2}{c|}{WER [\%]}  \\ 
\hline
Architecture                       & Data (hours)                                                                     & Epochs               & Hykist dev & Hykist test       \\ 
\hline
\multirow{2}{*}{\textit{Large}\textsubscript{1-8}} & None                                                                             & None                 & 27.6       & 31.9              \\ 
\cline{2-5}
                                   & \multirow{2}{*}{\begin{tabular}[c]{@{}c@{}}Viet. in-house \\(219h)\end{tabular}} & 25                   & 27.6       & 29.5              \\ 
\cline{1-1}\cline{3-5}
\textit{Large}                     &                                                                                  & 100                  & 26.2       & 29.0              \\ 
\hline
\multirow{3}{*}{\textit{Large}\textsubscript{1-8}} & \begin{tabular}[c]{@{}c@{}}Viet. YT \\(1168h)\end{tabular}                       & \multirow{2}{*}{100} & 24.3       & 28.1              \\ 
\cline{2-2}\cline{4-5}
                                   & \begin{tabular}[c]{@{}c@{}}Viet. in-house + YT~\\(1168h)\end{tabular}            &                      & 24.5       & 27.2              \\ 
\cline{2-5}
                                   & \begin{tabular}[c]{@{}c@{}}Multilingual in-house \\(1168h)\end{tabular}          & 50                   & 23.9       & 27.4              \\
\hline
\end{tabular}
\caption{\glspl{WER} {[}\%{]} for models using unsupervised pre-training with the public \textit{XLSR-53} model as initialization (3rd row is for direct finetuning and the rest are initialization for pretraining on specific data). All fine-tunings use the \textit{Large}\textsubscript{1-8} architecture and are trained until full convergence on Vietnamese in-house data and the recognition is done on HYKIST. The 1st model is the direct finetuning on Vietnamese in-house data, and the remaining models use \textit{XLSR-53} as initialization for pretrainings (full model \textit{Large} or cut-off model  \textit{Large}\textsubscript{1-8}).}
\label{table: xlsr53_init_pretrain}
\end{table}

As an alternative, we might use \textit{XLSR-53}\textsubscript{1-8} as an initialization for a customized pre-training, as shown in Table \ref{table: xlsr53_init_pretrain}. 
On the in-house Vietnamese data, the \glsxtrshort{WER}s reduce from 31.4\% and 33.4\% (Table \ref{table:monoling_pretraining}) to 27.6\% and 29.5\% on dev and test set respectively, compared to 27.6\% and 31.9\% of direct finetuning with \textit{XLSR-53}\textsubscript{1-8}.
This proves that continued pretraining using \textit{XLSR-53} model outperforms the pretraining using random initialization and the direct finetuning using \textit{XLSR-53}.

A \textit{Large} model initialized with \glsxtrshort{XLSR-53} is also pre-trained on the monolingual in-house data before being reduced to a smaller size for fine-tuning.
This performs better than pre-training with the smaller \textit{Large}\textsubscript{1-8} (26.2\% and 29.0\% compared to 27.6\% and 29.5\% on dev and test set respectively), but at the expense of increased pre-training's resource usage.
Therefore, if the resource usage is neglected, the \textit{Large} model should be chosen for better \glsxtrshort{WER}.

For the pretraining on the \glsxtrshort{YT} data using \glsxtrshort{XLSR-53} as initialization, the \glsxtrshort{WER}s reduce from 29.8\% and 35.2\% (Table \ref{table:monoling_pretraining}) to 24.3\% and 28.1\% on dev and test set respectively. 
The benefits of integrating \textit{XLSR-53} into the multilingual data are substantially lower, with \glsxtrshort{WER}s being reduced from 26.8\% and 28.7\% (Table \ref{table:multiling_pretraining}) to 23.9\% and 27.4\%.
On the domain-diverse dataset (the combination of monolingual in-house and \glsxtrshort{YT} data), the benefits of continued pretraining are also reduced, with \glsxtrshort{WER}s being reduced from 25.3\% and 27.2\% (Table \ref{table:monoling_pretraining}) to 24.5\% and 27.2\% on dev and test set respectively.
This shows that the continued pretraining is beneficial for both the monolingual and the multilingual scenario. However, the continued pretraining on less diverse data benefits more from the diverse and multilingual data.

\subsection{Comparison to supervised baselines}

\begin{table}[!ht]
\centering
\begin{tabular}{|c|c|c|c|c|} 
\hline
\multirow{2}{*}{AM}          & \multirow{2}{*}{Init}                & Pre-training                                                            & \multicolumn{2}{c|}{WER [\%]}  \\ 
\cline{3-5}
                             &                                      & Data (hours)                                                            & Hykist dev & Hykist test       \\ 
\hline
Transformer                  & \multirow{4}{*}{random}              & \multirow{2}{*}{None}                                                   & 31.0       & 35.1              \\ 
\cline{1-1}\cline{4-5}
\multirow{5}{*}{wav2vec 2.0} &                                      &                                                                         & 32.1       & 36.6              \\ 
\cline{3-5}
                             &                                      & \begin{tabular}[c]{@{}c@{}}Viet. in-house \\~(219h)\end{tabular}        & 31.4       & 33.4              \\ 
\cline{3-5}
                             &                                      & \begin{tabular}[c]{@{}c@{}}Viet. YT \\(1168h)\end{tabular}              & 29.8       & 35.2              \\ 
\cline{2-5}
                             & \multirow{2}{*}{\textit{XLSR-53}\textsubscript{1-8}} & \begin{tabular}[c]{@{}c@{}}Viet. in-house + YT \\(1168h)\end{tabular}   & 24.5       & 27.2              \\ 
\cline{3-5}
                             &                                      & \begin{tabular}[c]{@{}c@{}}Multilingual in-house~\\(1168h)\end{tabular} & 23.9       & 27.4              \\
\hline
\end{tabular}
\caption{\glspl{WER} {[}\%{]} for models using unsupervised pre-training and supervised-only training. All fine-tunings use the \textit{Large}\textsubscript{1-8} architecture and are trained until full convergence on Vietnamese in-house data and the recognition is done on HYKIST. The 1st model is the supervised-only training using Transformer. The 2nd and 3rd models are pretrained on specific data using random initializaton. The 4th and 5th models are continued pretraining methods (using \textit{XLSR-53}\textsubscript{1-8} as initialization).}
\label{table: supervised_unsupervised_compare}
\end{table}

As shown in Table \ref{table: supervised_unsupervised_compare}, we can see that fine-tuning using \glsxtrshort{Wav2vec 2.0} \textit{Large}\textsubscript{1-8}\xspace from scratch is worse when we compare with the findings from the supervised-only baseline (32.1\% and 36.6\% vs. 31.0\% and 35.1\% on dev and test set respectively).
With monolingual pre-training on the identical data, there is still no apparent advantage (31.4\% and 33.4\%).
When we increase the pretraining data to 5 times with a less diverse data (\glsxtrshort{YT} data), the performance also does not clearly outperform the supervised-only baseline (29.8\% and 35.2\%).
This proves that the \glsxtrshort{Wav2vec 2.0} unsupervised pretraining does not always outperform the \glsxtrshort{Transformer} supervised-only approach, especially when the pretrained data is not diverse enough.

However, we are able to significantly outperform the supervised baselines when applying continued pretraining.
In comparison to the best supervised-only baseline, the best results for continued pretraining show a reduction of \glspl{WER} to 24.5 \% and 27.2\% on monolingual data and to 23.9\% and 27.4\% on multilingual data.
Therefore, we can conclude that continued pretraining should be used to gain the most benefits in terms of accuracy.
    \section{Encoder and initialization comparison}

\subsection{Encoder comparison}

\begin{table}[!h]
\centering
\begin{adjustbox}{width=0.9\columnwidth,center}
\begin{tabular}{|c|c|c|c|} 
\hline
\multirow{2}{*}{Architecture} & Pretraining                                                                              & \multicolumn{2}{c|}{WER [\%]}  \\ 
\cline{2-4}
                              & Data (hours)                                                                             & Hykist dev & Hykist test       \\ 
\hline
\textit{Base}                 & \multirow{2}{*}{None}                                                                    & 35.8       & 39.9              \\ 
\cline{1-1}\cline{3-4}
\textit{Large}\textsubscript{1-8}             &                                                                                          & 35.0       & 40.7              \\ 
\hline
\textit{Base}                 & \multirow{2}{*}{\begin{tabular}[c]{@{}c@{}}Viet. in-house\\(219h)\end{tabular}}          & 30.2       & 33.3              \\ 
\cline{1-1}\cline{3-4}
\textit{Large}\textsubscript{1-8}             &                                                                                          & 31.5       & 33.4              \\ 
\hline
\textit{Base}                 & \multirow{2}{*}{\begin{tabular}[c]{@{}c@{}}Multilingual in-house~\\(1168h)\end{tabular}} & 26.2       & 28.8              \\ 
\cline{1-1}\cline{3-4}
\textit{Large}\textsubscript{1-8}             &                                                                                          & 26.8       & 28.7              \\
\hline
\end{tabular}
\end{adjustbox}
\caption{\glspl{WER} {[}\%{]} for architecture \textit{Base} and \textit{Large}\textsubscript{1-8} using different pretraining schedules: no pretraining, pretraining on in-house data and on multilingual data. All fine-tunings are done until full convergence on Vietnamese in-house data and the recognition is done on HYKIST.}
\label{table:encoder_compare_pretrain}
\end{table}

We compare the performance of 2 types of encoder: \textit{Base} and \textit{Large}\textsubscript{1-8}.
As shown in Table \ref{table:encoder_compare_pretrain}, we receive mix results for various pretraining schedules: no pretraining, pretraining on in-house data and pretraining on multilingual data.
It is mentioned by \cite{irie2019language} in language modeling that the \textit{Base} architecture works better than the \textit{Large}.
However, in acoustic modeling in \glsxtrshort{ASR}, our results prove against this statement.
Considering the amount of parameters between \textit{Base} and \textit{Large}\textsubscript{1-8}, 97M vs. 118M, we recommend the use of \textit{Base} in order to keep the performance competitive to \textit{Large}\textsubscript{1-8} while reducing the number of trainable parameters.

\subsection{Initialization comparison}

\begin{table}[!h]
\centering
\begin{adjustbox}{width=\columnwidth,center}
\begin{tabular}{|c|c|c|c|c|c|} 
\hline
\multirow{2}{*}{Architecture} & \multirow{2}{*}{Init. scheme}                                                     & \multicolumn{2}{c|}{Pretraining}                                                                          & \multicolumn{2}{c|}{WER [\%]}  \\ 
\cline{3-6}
                              &                                                                                   & Data (hours)                                                                      & Epochs                & Hykist dev & Hykist test       \\ 
\hline
\textit{Base}                 & \multirow{2}{*}{\begin{tabular}[c]{@{}c@{}}Kaiming Init.\\(Fairseq)\end{tabular}} & \multirow{2}{*}{\begin{tabular}[c]{@{}c@{}}Viet. in-house~\\(0.01h)\end{tabular}} & \multirow{2}{*}{1}    & 30.6       & 35.2              \\ 
\cline{1-1}\cline{5-6}
\textit{Large}\textsubscript{1-8}             &                                                                                   &                                                                                   &                       & 31.8       & 35.7              \\ 
\hline
\textit{Base}                 & \multirow{2}{*}{\begin{tabular}[c]{@{}c@{}}Glorot Init.\\(RETURNN)\end{tabular}}  & \multirow{2}{*}{None}                                                             & \multirow{2}{*}{None} & 35.8       & 39.9              \\ 
\cline{1-1}\cline{5-6}
\textit{Large}\textsubscript{1-8}             &                                                                                   &                                                                                   &                       & 35.0       & 40.7              \\
\hline
\end{tabular}
\end{adjustbox}
\caption{\glspl{WER} {[}\%{]} for architecture \textit{Base} and \textit{Large}\textsubscript{1-8} using 2 different initialization schemes: Kaiming Initialization (in Fairseq framework \cite{facebook2019fairseq}) and Glorot Initialization (in RETURNN framework \cite{doetsch2016returnn}). All fine-tunings are done until full convergence on Vietnamese in-house data and the recognition is done on HYKIST.}
\label{table:encoder_compare_shortPretrain}
\end{table}

In the case of super short pretraining (1 epoch pretraining on only 0.01h of data), the results outperform those of raw waveform from scratch for both \textit{Base} and \textit{Large}\textsubscript{1-8} architecture as shown in Table \ref{table:encoder_compare_shortPretrain}.
The reason for the improvement comes from the difference of initialization schemes.
The parameters from the pretrained model are first initialized by Fairseq \cite{facebook2019fairseq} using Kaiming Initialization \cite{He_2015_ICCV}, and then fed into RETURNN \cite{doetsch2016returnn}, while the parameters for raw waveform training are initialized directly by RETURNN using Glorot (also known as Xavier) Initialization \cite{glorot2010understanding}.
We therefore recommend the use of Kaiming Initialization for \glsxtrshort{Wav2vec 2.0} architecture.

    \section{Effectiveness of intermediate loss}

\subsection{Effectiveness of Intermediate Cross-Entropy Loss}

\textbf{Improvement on HYKIST data}: In Table \ref{table:int_loss_hykist_pos}, when using in-house telephone dataset to train and transcribe the HYKIST dataset with the help of \glsxtrshort{ICE Loss}, we report the total improvement in performance for from scratch experiment where the \Glspl{WER} decrease from 35.6\% and 40.7\% to 33.8\% and 38.1\% on dev and test set respectively.
For \glsxtrshort{YT} experiment, the \Glspl{WER} decrease from 29.8\% and 35.2\% to 27.3\% and 31.5\%.
We also report a small improvement for the combination of Vietnamese in-house data and \glsxtrshort{YT} data, from 25.3\% and 27.2\% to 25.1\% and 27.1\%.

\begin{table}[!ht]
\centering
\begin{adjustbox}{width=0.9\columnwidth,center}
\begin{tabular}{|c|c|c|c|} 
\hline
\multirow{2}{*}{Pre-training data}                                                    & \multirow{2}{*}{With ICE} & \multicolumn{2}{c|}{WER [\%]}  \\ 
\cline{3-4}
                                                                                      &                           & Hykist dev & Hykist test       \\ 
\hline
\multirow{2}{*}{None}                                                                 & No                        & 35.6       & 40.7              \\ 
\cline{2-4}
                                                                                      & Yes                       & 33.8       & 38.1              \\ 
\hline
\multirow{2}{*}{\begin{tabular}[c]{@{}c@{}}Viet. YT\\(1168h)\end{tabular}}            & No                        & 29.8       & 35.2              \\ 
\cline{2-4}
                                                                                      & Yes                       & 27.3       & 31.5              \\ 
\hline
\multirow{2}{*}{\begin{tabular}[c]{@{}c@{}}Viet. in-house + YT\\(1168h)\end{tabular}} & No                        & 25.3       & 27.2              \\ 
\cline{2-4}
                                                                                      & Yes                       & 25.1       & 27.1              \\
\hline
\end{tabular}
\end{adjustbox}
\caption{
    Improvements of \glspl{WER} {[}\%{]} on HYKIST data between pretraining schedules when applying \glsxtrshort{ICE Loss}. All models are finetuned until full convergence on Vietnamese in-house data. 
    Only 1 intermediate layer is applied in the middle \glsxtrshort{Transformer} block, e.g. position 4 for \textit{Large}\textsubscript{1-8} and 6 for \textit{Base} architecture.}
\label{table:int_loss_hykist_pos}
\end{table}

\textbf{Degradation on HYKIST data}: As shown in Table \ref{table:int_loss_hykist_neg}, for the directly finetuning experiment with \glsxtrshort{XLSR-53} preloaded, the performance is hurt totally (both \Glspl{WER} on dev and test sets increase).
Besides, both continued pretrainings on Vietnamese in-house and on \glsxtrshort{YT} data experience the partial improvements (only \Glspl{WER} on test sets are slightly increased but \Glspl{WER} on dev sets decrease).
The rest pretraining schedules in Table \ref{table:int_loss_hykist_neg} also experience partial improvements.

\begin{table}[!ht]
\centering
\begin{adjustbox}{width=\columnwidth,center}
\begin{tabular}{|c|c|c|c|c|c|} 
\hline
\multirow{2}{*}{Arch.}         & \multirow{2}{*}{Init.}             & \multirow{2}{*}{Pre-training data}                                                    & \multirow{2}{*}{With ICE} & \multicolumn{2}{c|}{WER [\%]}  \\ 
\cline{5-6}
                               &                                    &                                                                                       &                           & Hykist dev & Hykist test       \\ 
\hline
\multirow{6}{*}{\textit{Large}\textsubscript{1-8}}      & \multirow{2}{*}{\textit{XLSR-53}}  & \multirow{2}{*}{None}                                                                 & No                        & 27.9       & 32.3              \\ 
\cline{4-6}
                               &                                    &                                                                                       & Yes                       & 28.4       & 33.3              \\ 
\cline{2-6}
                               & \multirow{2}{*}{None}              & \multirow{6}{*}{\begin{tabular}[c]{@{}c@{}}Viet. in-house\\(219h)\end{tabular}}       & No                        & 30.4       & 33.4              \\ 
\cline{4-6}
                               &                                    &                                                                                       & Yes                       & 29.1       & 33.7              \\ 
\cline{2-2}\cline{4-6}
                               & \multirow{2}{*}{\textit{XLSR-53 }} &                                                                                       & No                        & 25.5       & 29.1              \\ 
\cline{4-6}
                               &                                    &                                                                                       & Yes                       & 25.2       & 29.2              \\ 
\cline{1-2}\cline{4-6}
\multirow{2}{*}{\textit{Base}} & \multirow{4}{*}{None}              &                                                                                       & No                        & 30.2       & 33.3              \\ 
\cline{4-6}
                               &                                    &                                                                                       & Yes                       & 29.7       & 33.4              \\ 
\cline{1-1}\cline{3-6}
\multirow{4}{*}{\textit{Large}\textsubscript{1-8}}      &                                    & \multirow{2}{*}{\begin{tabular}[c]{@{}c@{}}Multiling. in-house\\(1168h)\end{tabular}} & No                        & 26.8       & 28.7              \\ 
\cline{4-6}
                               &                                    &                                                                                       & Yes                       & 25.5       & 29.4              \\ 
\cline{2-6}
                               & \multirow{2}{*}{\textit{XLSR-53}}  & \multirow{2}{*}{\begin{tabular}[c]{@{}c@{}}Viet. YT\\(1168h)\end{tabular}}            & No                        & 24.3       & 28.1              \\ 
\cline{4-6}
                               &                                    &                                                                                       & Yes                       & 23.7       & 28.2              \\
\hline
\end{tabular}
\end{adjustbox}
\caption{
    Degradations of \glspl{WER} {[}\%{]} on HYKIST data between pretraining schedules when applying \glsxtrshort{ICE Loss}. All models are finetuned until full convergence on Vietnamese in-house data. 
    Only 1 intermediate layer is applied in the middle \glsxtrshort{Transformer} block, e.g. position 4 for \textit{Large}\textsubscript{1-8} and 6 for \textit{Base} architecture.
    }
\label{table:int_loss_hykist_neg}
\end{table}

\textbf{Improvement on CommonVoice and VIVOS data}: In the situation of more out-of-domain recognition shown in Table \ref{int_loss_cvvivos_pos}, which means using the model finetuned on our in-house spontaneous telephone speech dataset to do the recognition on read speech datasets like CommonVoice and VIVOS, we report the total improvements in performance for \textit{Large}\textsubscript{1-8} in-house pretraining, from scratch and \glsxtrshort{YT} experiments. 
Notable is from scratch training where \glsxtrshort{WER}s reduce from 20.8\%, 44.7\%, 34.9\% to 18.6\%, 42.1\%, 33.1\%; and \glsxtrshort{YT} pretraining where \glsxtrshort{WER}s reduce from 16.4\%, 34.4\%, 28.7\% to 15.6\%, 32.2\%, 27.6\% on CommonVoice dev/test and VIVOS test set respectively.
Together with the improvements on HYKIST reported in Table \ref{table:int_loss_hykist_pos}, we conclude that using \glsxtrshort{ICE Loss} for from scratch training and for pretraining on \glsxtrshort{YT} data improves the recognitions on both telephone and read speech domain.

\begin{table}[!ht]
\centering
\begin{adjustbox}{width=0.7\columnwidth,center}
\begin{tabular}{|c|c|c|c|c|} 
\hline
\multirow{2}{*}{Pre-training data}                                              & \multirow{2}{*}{With ICE} & \multicolumn{3}{c|}{WER [\%]}  \\ 
\cline{3-5}
                                                                                &                           & CV dev & CV test & Vivos       \\ 
\hline
\multirow{2}{*}{None}                                                           & No                        & 20.8   & 44.7    & 34.9        \\ 
\cline{2-5}
                                                                                & Yes                       & 18.6   & 42.1    & 33.1        \\ 
\hline
\multirow{2}{*}{\begin{tabular}[c]{@{}c@{}}Viet. YT\\(1168h)\end{tabular}}      & No                        & 16.4   & 34.4    & 28.7        \\ 
\cline{2-5}
                                                                                & Yes                       & 15.6   & 32.2    & 27.6        \\ 
\hline
\multirow{2}{*}{\begin{tabular}[c]{@{}c@{}}Viet. in-house\\(219h)\end{tabular}} & No                        & 16.4   & 35.6    & 31.3        \\ 
\cline{2-5}
                                                                                & Yes                       & 16.1   & 34.8    & 30.4        \\
\hline
\end{tabular}
\end{adjustbox}
\caption{
    Improvements of \glspl{WER} {[}\%{]} on CommonVoice and VIVOS between pretraining schedules when applying \glsxtrshort{ICE Loss}. All models are finetuned on Vietnamese in-house data. 
    Only 1 intermediate layer is applied in the middle \glsxtrshort{Transformer} block, e.g. position 4 for \textit{Large}\textsubscript{1-8} and 6 for \textit{Base} architecture.
    }
\label{int_loss_cvvivos_pos}
\end{table}

\textbf{Degradation on CommonVoice and VIVOS data}: As shown in the Table \ref{int_loss_cvvivos_neg}, we experience the total degradations for 2 cases: continued pretraining on Vietnamese in-house data and pretraining on the combination of in-house and \glsxtrshort{YT} data; where \glsxtrshort{WER}s for all read speech test sets increase.
The rest cases experience partial degradations.

\begin{table}[!ht]
\centering
\begin{adjustbox}{width=\columnwidth,center}
\begin{tabular}{|c|c|c|c|c|c|c|} 
\hline
\multirow{2}{*}{Arch.}             & \multirow{2}{*}{Init.}            & \multirow{2}{*}{Pre-training data}                                                    & \multirow{2}{*}{With ICE} & \multicolumn{3}{c|}{WER [\%]}  \\ 
\cline{5-7}
                                   &                                   &                                                                                       &                           & CV dev & CV test & Vivos       \\ 
\hline
\multirow{4}{*}{\textit{Large}\textsubscript{1-8}} & \multirow{4}{*}{\textit{XLSR-53}} & \multirow{2}{*}{None}                                                                 & No                        & 14.8   & 32.5    & 30.3        \\ 
\cline{4-7}
                                   &                                   &                                                                                       & Yes                       & 15.8   & 33.9    & 30.0        \\ 
\cline{3-7}
                                   &                                   & \multirow{4}{*}{\begin{tabular}[c]{@{}c@{}}Viet. in-house\\(219h)\end{tabular}}       & No                        & 11.5   & 29.4    & 27.2        \\ 
\cline{4-7}
                                   &                                   &                                                                                       & Yes                       & 12.3   & 29.8    & 27.7        \\ 
\cline{1-2}\cline{4-7}
\multirow{2}{*}{\textit{Base}}     & \multirow{6}{*}{None}             &                                                                                       & No                        & 16.6   & 35.4    & 30.9        \\ 
\cline{4-7}
                                   &                                   &                                                                                       & Yes                       & 15.4   & 34.2    & 31.3        \\ 
\cline{1-1}\cline{3-7}
\multirow{6}{*}{\textit{Large}\textsubscript{1-8}} &                                   & \multirow{2}{*}{\begin{tabular}[c]{@{}c@{}}Multiling. in-house\\(1168h)\end{tabular}} & No                        & 15.2   & 29.7    & 29.5        \\ 
\cline{4-7}
                                   &                                   &                                                                                       & Yes                       & 14.8   & 30.5    & 28.8        \\ 
\cline{3-7}
                                   &                                   & \multirow{2}{*}{\begin{tabular}[c]{@{}c@{}}Viet. in-house + YT\\(1168h)\end{tabular}} & No                        & 12.9   & 26.5    & 21.0        \\ 
\cline{4-7}
                                   &                                   &                                                                                       & Yes                       & 13.6   & 28.2    & 21.9        \\ 
\cline{2-7}
                                   & \multirow{2}{*}{\textit{XLSR-53}} & \multirow{2}{*}{\begin{tabular}[c]{@{}c@{}}Viet. YT\\(1168h)\end{tabular}}            & No                        & 11.8   & 28.4    & 25.6        \\ 
\cline{4-7}
                                   &                                   &                                                                                       & Yes                       & 12.3   & 28.3    & 25.0        \\
\hline
\end{tabular}
\end{adjustbox}
\caption{
    Degradations of \glspl{WER} {[}\%{]} on CommonVoice and VIVOS between pretraining schedules when applying \glsxtrshort{ICE Loss}. All models are finetuned on Vietnamese in-house data. 
    Only 1 intermediate layer is applied in the middle \glsxtrshort{Transformer} block, e.g. position 4 for \textit{Large}\textsubscript{1-8} and 6 for \textit{Base} architecture.}
\label{int_loss_cvvivos_neg}
\end{table}

\bigskip

\subsection{Effectiveness of Intermediate Focal Loss}

\textbf{Effectiveness on HYKIST data}: 

As shown in Table \ref{table:if_loss_hykist_pos} and Table \ref{table:if_loss_hykist_neg} below, when using \glsxtrshort{IF Loss}, we see the \Glspl{WER} on HYKIST improved compared to the baselines for various pretraining schedules (7/9 experiments experience total improvements), compared to only 3/9 experiments experiencing total improvements using \glsxtrshort{ICE Loss} (\glsxtrshort{ICE Loss} results are shown in Table \ref{table:int_loss_hykist_pos} and Table \ref{table:int_loss_hykist_neg} above).
In addition, we report all \Glspl{WER} of \glsxtrshort{IF Loss} experiments to be lower than those of \glsxtrshort{ICE Loss} experiments, except the one on HYKIST test set of from scratch training. 
We therefore conclude that, when finetuning and recognizing on the same telephone domain, \glsxtrshort{IF Loss} works better than \glsxtrshort{ICE Loss}. 

Compared to our strongest continued pretraining baseline, the application of \glsxtrshort{IF Loss} on the combination of Vietnamese in-house and \glsxtrshort{YT} data (24.5\% and 27.1\%) outperforms the results of continued pretraining on the combination of Vietnamese in-house and \glsxtrshort{YT} data (24.5\% and 27.2\% on dev and test set respectively as shown in Table \ref{table: xlsr53_init_pretrain}).
However, we believe that the \glsxtrshort{IF Loss} can further reduce \glsxtrshort{WER}s for this continued pretraining schedule, as it does with continued pretraining on \glsxtrshort{YT} data. 

Among all total improvements reported in Table \ref{table:if_loss_hykist_pos}, notable is the \Glspl{WER} reduction of \glsxtrshort{YT} experiment from 29.8\% and 35.2\% to 26.1\% and 30.8\% on dev and test set respectively, whose relative \glsxtrshort{WERR} is around 12.5\% in average.
For a more diverse pretrained data (Vietnamese in-house data), we report the \glsxtrshort{WER}s reduction from 30.4\% and 33.4\% to 28.6\% and 33.0\%, whose relative \glsxtrshort{WERR} is around 3.6\% in average.
For even more diverse pretrained data (Vietnamese in-house  + \glsxtrshort{YT} data), we report the \glsxtrshort{WER}s reduction from 25.3\% and 27.2\% to 24.5\% and 27.1\%, whose relative \glsxtrshort{WERR} is around 1.8\% in average.
We therefore conclude that the effectiveness of \glsxtrshort{IF Loss} decreases when the pretrained data becomes more diverse.

\begin{table}[!ht]
\centering
\begin{adjustbox}{width=\columnwidth,center}
\begin{tabular}{|c|c|c|c|c|c|} 
\hline
\multirow{2}{*}{Arch.}             & \multirow{2}{*}{Init.}             & \multirow{2}{*}{Pre-training data}                                                    & \multirow{2}{*}{With IF} & \multicolumn{2}{c|}{WER [\%]}  \\ 
\cline{5-6}
                                   &                                    &                                                                                       &                          & Hykist dev & Hykist test       \\ 
\hline
\multirow{6}{*}{\textit{Large}\textsubscript{1-8}} & \multirow{4}{*}{None}              & \multirow{2}{*}{None}                                                                 & No                       & 35.6       & 40.7              \\ 
\cline{4-6}
                                   &                                    &                                                                                       & Yes                      & 33.0       & 38.8              \\ 
\cline{3-6}
                                   &                                    & \multirow{6}{*}{\begin{tabular}[c]{@{}c@{}}Viet. in-house\\(219h)\end{tabular}}       & No                       & 30.4       & 33.4              \\ 
\cline{4-6}
                                   &                                    &                                                                                       & Yes                      & 28.6       & 33.0              \\ 
\cline{2-2}\cline{4-6}
                                   & \multirow{2}{*}{\textit{XLSR-53 }} &                                                                                       & No                       & 25.5       & 29.1              \\ 
\cline{4-6}
                                   &                                    &                                                                                       & Yes                      & 24.7       & 29.1              \\ 
\cline{1-2}\cline{4-6}
\multirow{2}{*}{\textit{Base}}     & \multirow{6}{*}{None}              &                                                                                       & No                       & 30.2       & 33.3              \\ 
\cline{4-6}
                                   &                                    &                                                                                       & Yes                      & 29.0       & 33.0              \\ 
\cline{1-1}\cline{3-6}
\multirow{6}{*}{\textit{Large}\textsubscript{1-8}} &                                    & \multirow{2}{*}{\begin{tabular}[c]{@{}c@{}}Viet. YT\\(1168h)\end{tabular}}            & No                       & 29.8       & 35.2              \\ 
\cline{4-6}
                                   &                                    &                                                                                       & Yes                      & 26.1       & 30.8              \\ 
\cline{3-6}
                                   &                                    & \multirow{2}{*}{\begin{tabular}[c]{@{}c@{}}Viet. in-house + YT\\(1168h)\end{tabular}} & No                       & 25.3       & 27.2              \\ 
\cline{4-6}
                                   &                                    &                                                                                       & Yes                      & 24.5       & 27.1              \\ 
\cline{2-6}
                                   & \multirow{2}{*}{\textit{XLSR-53}}  & \multirow{2}{*}{\begin{tabular}[c]{@{}c@{}}Viet. YT\\(1168h)\end{tabular}}            & No                       & 24.3       & 28.1              \\ 
\cline{4-6}
                                   &                                    &                                                                                       & Yes                      & 23.4       & 28.1              \\
\hline
\end{tabular}
\end{adjustbox}
\caption{
    Improvements of \glspl{WER} {[}\%{]} on HYKIST data between pretraining schedules when applying \glsxtrshort{IF Loss}. All models are finetuned until full convergence on Vietnamese in-house data. 
    Only 1 intermediate layer is applied in the middle \glsxtrshort{Transformer} block, e.g. position 4 for \textit{Large}\textsubscript{1-8} and 6 for \textit{Base} architecture.}
\label{table:if_loss_hykist_pos}
\end{table}

As shown in Table \ref{table:if_loss_hykist_neg}, only for the case of directly finetuning with \glsxtrshort{XLSR-53}, using \glsxtrshort{IF Loss} makes the \Glspl{WER} on HYKIST increased compared to the baselines. 
However, the degradation is rather small, from 27.9\% and 32.3\% to 28.0\% and 32.8\% on dev and test set respectively.
Besides, a partial degradation of performance is reported in the multilingual in-house data experiment, where the  average \glsxtrshort{WER} of dev and test set (25.2\% and 29.3\%) is even lower than the baseline (26.8\% and 28.7\%).
Hence, in a rapid deployment of an \glsxtrshort{ASR} system, we recommend the direct use of \glsxtrshort{IF Loss} in training without the need of one more training as a baseline for performance comparison.

\begin{table}[!ht]
\centering
\begin{adjustbox}{width=0.9\columnwidth,center}
\begin{tabular}{|c|c|c|c|c|} 
\hline
\multirow{2}{*}{Init.}            & \multirow{2}{*}{Pre-training data}                                                    & \multirow{2}{*}{With IF} & \multicolumn{2}{c|}{WER [\%]}  \\ 
\cline{4-5}
                                  &                                                                                       &                          & Hykist dev & Hykist test       \\ 
\hline
\multirow{2}{*}{\textit{XLSR-53}} & \multirow{2}{*}{None}                                                                 & No                       & 27.9       & 32.3              \\ 
\cline{3-5}
                                  &                                                                                       & Yes                      & 28.0       & 32.8              \\ 
\hline
\multirow{2}{*}{None}             & \multirow{2}{*}{\begin{tabular}[c]{@{}c@{}}Multiling. in-house\\(1168h)\end{tabular}} & No                       & 26.8       & 28.7              \\ 
\cline{3-5}
                                  &                                                                                       & Yes                      & 25.2       & 29.3              \\
\hline
\end{tabular}
\end{adjustbox}
\caption{
    Degradations of \glspl{WER} {[}\%{]} on HYKIST data between pretraining schedules when applying \glsxtrshort{IF Loss}. All models are finetuned until full convergence on Vietnamese in-house data. 
    Only 1 intermediate layer is applied in the middle \glsxtrshort{Transformer} block, e.g. position 4 for \textit{Large}\textsubscript{1-8} and 6 for \textit{Base} architecture.}
\label{table:if_loss_hykist_neg}
\end{table}

\bigskip

\textbf{Effectiveness on CommonVoice and VIVOS data}: 

In the larger domain-shift recognition, we still receive the significant reduction of \Glspl{WER} in multiple experiments as shown in Table \ref{table:if_loss_cvvivos_pos}. 
The notable reduction of \Glspl{WER} compared to baselines is again on \glsxtrshort{YT} data, whose \Glspl{WER}s decrease from 16.4\%, 34.4\% and 28.7\% to 14.5\%, 30.9\% and 26.9\% respectively for 3 read speech sets, that makes \glsxtrshort{WERR} about 9.3\% in average.
The \glsxtrshort{ICE Loss} in Table \ref{int_loss_cvvivos_pos} makes 3 experiments totally improved, while the \glsxtrshort{IF Loss} makes 4. 
Furthermore, the \glsxtrshort{WER}s for \glsxtrshort{IF Loss} on 3 read speech datasets are as competitive as \glsxtrshort{ICE Loss}.
In addition, when finetuning and recognizing on the same telephone domain, \glsxtrshort{IF Loss} works better than \glsxtrshort{ICE Loss} as proved above.
We therefore conclude that \glsxtrshort{IF Loss} works better than \glsxtrshort{ICE Loss} in all domains.

\begin{table}[!ht]
\centering
\begin{adjustbox}{width=0.9\columnwidth,center}
\begin{tabular}{|c|c|c|c|c|c|} 
\hline
\multirow{2}{*}{Arch.}             & \multirow{2}{*}{Pre-training data}                                              & \multirow{2}{*}{With IF} & \multicolumn{3}{c|}{WER [\%]}  \\ 
\cline{4-6}
                                   &                                                                                 &                          & CV dev & CV test & Vivos       \\ 
\hline
\multirow{4}{*}{\textit{Large}\textsubscript{1-8}} & \multirow{2}{*}{\begin{tabular}[c]{@{}c@{}}Viet. in-house\\(219h)\end{tabular}} & No                       & 16.4   & 35.6    & 31.3        \\ 
\cline{3-6}
                                   &                                                                                 & Yes                      & 15.8   & 34.5    & 29.6        \\ 
\cline{2-6}
                                   & \multirow{2}{*}{None}                                                           & No                       & 20.8   & 44.7    & 34.9        \\ 
\cline{3-6}
                                   &                                                                                 & Yes                      & 19.7   & 43.1    & 33.9        \\ 
\hline
\multirow{2}{*}{\textit{Base}}     & \multirow{2}{*}{\begin{tabular}[c]{@{}c@{}}Viet. in-house\\(219h)\end{tabular}} & No                       & 16.6   & 35.4    & 30.9        \\ 
\cline{3-6}
                                   &                                                                                 & Yes                      & 15.9   & 34.4    & 30.5        \\ 
\hline
\multirow{2}{*}{\textit{Large}\textsubscript{1-8}} & \multirow{2}{*}{\begin{tabular}[c]{@{}c@{}}Viet. YT\\(1168h)\end{tabular}}      & No                       & 16.4   & 34.4    & 28.7        \\ 
\cline{3-6}
                                   &                                                                                 & Yes                      & 14.5   & 30.9    & 26.9        \\
\hline
\end{tabular}
\end{adjustbox}
\caption{
    Improvements of \glspl{WER} {[}\%{]} on CommonVoice and VIVOS between pretraining schedules when applying \glsxtrshort{IF Loss}. All models are finetuned on Vietnamese in-house data. 
    Only 1 intermediate layer is applied in the middle \glsxtrshort{Transformer} block, e.g. position 4 for \textit{Large}\textsubscript{1-8} and 6 for \textit{Base} architecture.}
\label{table:if_loss_cvvivos_pos}
\end{table}

However, in the larger domain-shift recognition, we still meet degradations of performance in experiments pretrained on diverse data, as shown in Table \ref{table:if_loss_cvvivos_neg}.
We therefore recommend the use of \glsxtrshort{IF Loss} only for less diverse pretrained data if the domain of finetuning and recognition data are too different.

\begin{table}[!ht]
\centering
\begin{adjustbox}{width=0.9\columnwidth,center}
\begin{tabular}{|c|c|c|c|c|c|} 
\hline
\multirow{2}{*}{Init.}            & \multirow{2}{*}{Pre-training data}                                                    & \multirow{2}{*}{With IF} & \multicolumn{3}{c|}{WER [\%]}  \\ 
\cline{4-6}
                                  &                                                                                       &                          & CV dev & CV test & Vivos       \\ 
\hline
\multirow{4}{*}{\textit{XLSR-53}} & \multirow{2}{*}{None}                                                                 & No                       & 14.8   & 32.5    & 30.3        \\ 
\cline{3-6}
                                  &                                                                                       & Yes                      & 15.4   & 33.6    & 30.0        \\ 
\cline{2-6}
                                  & \multirow{2}{*}{\begin{tabular}[c]{@{}c@{}}Viet. in-house\\(219h)\end{tabular}}       & No                       & 11.5   & 29.4    & 27.2        \\ 
\cline{3-6}
                                  &                                                                                       & Yes                      & 13.0   & 29.8    & 27.6        \\ 
\hline
\multirow{4}{*}{None}             & \multirow{2}{*}{\begin{tabular}[c]{@{}c@{}}Multiling. in-house\\(1168h)\end{tabular}} & No                       & 15.2   & 29.7    & 29.5        \\ 
\cline{3-6}
                                  &                                                                                       & Yes                      & 14.5   & 30.6    & 28.1        \\ 
\cline{2-6}
                                  & \multirow{2}{*}{\begin{tabular}[c]{@{}c@{}}Viet. in-house + YT\\(1168h)\end{tabular}} & No                       & 12.9   & 26.5    & 21.0        \\ 
\cline{3-6}
                                  &                                                                                       & Yes                      & 12.7   & 28.6    & 22.1        \\ 
\hline
\multirow{2}{*}{\textit{XLSR-53}} & \multirow{2}{*}{\begin{tabular}[c]{@{}c@{}}Viet. YT\\(1168h)\end{tabular}}            & No                       & 11.8   & 28.4    & 25.6        \\ 
\cline{3-6}
                                  &                                                                                       & Yes                      & 13.2   & 29.1    & 24.5        \\
\hline
\end{tabular}
\end{adjustbox}
\caption{
    Degradations of \glspl{WER} {[}\%{]} on CommonVoice and VIVOS between pretraining schedules when applying \glsxtrshort{IF Loss}. All models are finetuned on Vietnamese in-house data. 
    Only 1 intermediate layer is applied in the middle \glsxtrshort{Transformer} block, e.g. position 4 for \textit{Large}\textsubscript{1-8} and 6 for \textit{Base} architecture.}
\label{table:if_loss_cvvivos_neg}
\end{table}

    \section{Intermediate loss analysis}

\subsection{Studies on Intermediate Focal Loss design}

\begin{table}[!ht]
\centering
\begin{adjustbox}{width=0.8\columnwidth,center}
\begin{tabular}{|c|c|c|c|c|c|} 
\hline
\multicolumn{6}{|c|}{Viet. in-house~\textit{Large}\textsubscript{1-8}}                                                                              \\ 
\hline
Layer & Hykist dev             & Hykist test            & CV dev                 & CV test                & Vivos                   \\ 
\hline
None  & 30.4                   & 33.4                   & 16.4                   & 35.6                   & 31.3                    \\ 
\hline
2     & 29.4                   & 34.0                   & \textbf{15.5}          & 35.7                   & 30.0                    \\ 
\hline
4     & \textbf{\textbf{28.6}} & \textbf{\textbf{33.0}} & 15.8                   & \textbf{\textbf{34.5}} & \textbf{\textbf{29.6}}  \\ 
\hline
6     & 29.1                   & \textbf{33.0}          & 16.6                   & 34.1                   & 29.9                    \\ 
\hline
2,6   & 29.1                   & 34.1                   & 15.6                   & 35.5                   & 30.7                    \\ 
\hline
3,5   & 29.1                   & 33.3                   & 16.4                   & 35.0                   & 30.1                    \\ 
\hline
\multicolumn{6}{|c|}{Viet. in-house~\textit{Base}}                                                                                  \\ 
\hline
None  & 30.2                   & 33.3                   & 16.6                   & 35.4                   & 30.9                    \\ 
\hline
3     & \textbf{28.7}          & 33.4                   & 17.0                   & 35.5                   & \textbf{30.1}           \\ 
\hline
6     & 29.0                   & 33.0                   & \textbf{\textbf{15.9}} & \textbf{\textbf{34.4}} & 30.5                    \\ 
\hline
9     & 29.5                   & \textbf{32.6}          & 14.8                   & 34.4                   & 30.5                    \\ 
\hline
4,8   & 29.3                   & 33.5                   & 16.2                   & 35.0                   & 30.3                    \\
\hline
\end{tabular}
\end{adjustbox}
\caption{
    \glspl{WER} {[}\%{]} comparison of \glsxtrshort{IF Loss} on different layers between 2 architecture sizes: \textit{Base} and \textit{Large}\textsubscript{1-8}.
    All models are finetuned until full convergence on Vietnamese in-house data and recognized on HYKIST, CommonVoice and VIVOS dataset.
    Layer "None" means the baseline (no application of \glsxtrshort{IF Loss}).
    }
\label{WER_intLoss_multiple_layer}
\end{table}

In Table \ref{WER_intLoss_multiple_layer}, we study variants of putting \glsxtrshort{IF Loss} at different layers. 
For \textit{Large}\textsubscript{1-8} model, we observe the performance degradation when moving the single \glsxtrshort{IF Loss} to different layers, while this gives mix results for the \textit{Base} model.
\cite{lee2021intermediate} also reports the same behavior when using Intermediate CTC Loss on 12-layer, 24-layer and 48-layer models in a supervised-only scenario. 
When applying 2 intermediate layers, we meet the degradation of performance for both \textit{Base} and \textit{Large}\textsubscript{1-8} models. 
From the experimental results, we therefore conclude that: Single \glsxtrshort{IF Loss} in the middle network layer yields the best result among variants.

\subsection{On-off Regularization technique}

\begin{table}[!ht]
\centering
\begin{adjustbox}{width=\columnwidth,center}
\begin{tabular}{|c|c|c|c|c|c|c|} 
\hline
\multicolumn{3}{|c|}{Off reg.}          & \multirow{6}{*}{\begin{tabular}[c]{@{}c@{}}Continue fine-tuning\\$\longrightarrow$\\\end{tabular}} & \multicolumn{3}{c|}{On reg.}             \\ 
\cline{1-3}\cline{5-7}
~Epochs & Hykist dev    & Hykist test   &                                                                                       & ~Epochs & Hykist dev    & Hykist test    \\ 
\cline{1-3}\cline{5-7}
0       & -             & -             &                                                                                       & 33      & 33.0          & 38.8           \\ 
\cline{1-3}\cline{5-7}
3       & 45.6          & 48.6          &                                                                                       & 33      & \textbf{32.5} & \textbf{38.4}  \\ 
\cline{1-3}\cline{5-7}
7       & 43.4 & 45.9 &                                                                                       & 33      & 34.3          & 39.5           \\ 
\cline{1-3}\cline{5-7}
10      & 44.6          & 47.1          &                                                                                       & 33      & 34.1          & 39.5           \\
\hline
\end{tabular}
\end{adjustbox}
\caption{
    \glspl{WER} {[}\%{]} comparison of On-off Regularization technique over epochs for raw waveform from scratch training.
    "Off Regularization" stage means training without regularization techniques like Dropout, SpecAugment and \glsxtrshort{IF Loss}.
    After training for some first epochs, the learning rate is reset and the model is preloaded in the "On Regularization" stage (all regularization techniques are turned on).
    All models are then continued being finetuned until full convergence on Vietnamese in-house data and recognized on HYKIST dataset.
    The 3rd row (0 epoch for "Off Regularization") is the baseline.}
\label{on_off_regularization}
\end{table}

To better exploit the \glsxtrshort{IF Loss}, we introduce the "\textbf{On-off Regularization} technique". 
We experiment this technique for raw waveform from scratch training.
Experimental results in Table \ref{on_off_regularization} show that, if we train without any regularization techniques ("Off Regularization" stage) for the first 3 epochs and then reset the learning rate and continue training with all regularizations turned on ("On Regularization" stage), we achieve the \glsxtrshort{WER}s reduction from 33.0\% and 38.8\% to 32.5\% and 38.4\% on dev and test set respectively compared to the baseline. 

In the future work, we plan to apply the "On-off Regularization" technique to other pretraining schedules.

\subsection{Combination of L2 regularization and Intermediate Focal Loss}

\begin{table}[!ht]
\centering
\begin{adjustbox}{width=\columnwidth,center}
\begin{tabular}{|c|c|c|c|c|c|} 
\hline
\multirow{2}{*}{Arch.}             & \multirow{2}{*}{Init.}            & \multirow{2}{*}{Pre-training data}                                              & \multirow{2}{*}{Reg.} & \multicolumn{2}{c|}{WER [\%]}  \\ 
\cline{5-6}
                                   &                                   &                                                                                 &                       & Hykist dev & Hykist test       \\ 
\hline
\multirow{4}{*}{\textit{Large}\textsubscript{1-8}} & \multirow{6}{*}{None}             & \multirow{2}{*}{\begin{tabular}[c]{@{}c@{}}Viet. in-house\\(219h)\end{tabular}} & With IF               & 28.6       & 33.0              \\ 
\cline{4-6}
                                   &                                   &                                                                                 & With IF + L2          & 28.6       & 32.9              \\ 
\cline{3-6}
                                   &                                   & \multirow{2}{*}{None}                                                           & With IF               & 33.0       & 38.8              \\ 
\cline{4-6}
                                   &                                   &                                                                                 & With IF + L2          & 31.4       & 36.4              \\ 
\cline{1-1}\cline{3-6}
\multirow{2}{*}{\textit{Base}}     &                                   & \multirow{4}{*}{\begin{tabular}[c]{@{}c@{}}Viet. in-house\\(219h)\end{tabular}} & With IF               & 29.0       & 33.0              \\ 
\cline{4-6}
                                   &                                   &                                                                                 & With IF + L2          & 28.8       & 32.3              \\ 
\cline{1-2}\cline{4-6}
\multirow{2}{*}{\textit{Large}\textsubscript{1-8}} & \multirow{2}{*}{\textit{XLSR-53}} &                                                                                 & With IF               & 24.7       & 29.1              \\ 
\cline{4-6}
                                   &                                   &                                                                                 & With IF + L2          & 24.3       & 29.2              \\
\hline
\end{tabular}
\end{adjustbox}
\caption{
    \glspl{WER} {[}\%{]} comparison of the L2 Regularization combination with \glsxtrshort{IF Loss}.
    All models are finetuned until full convergence on Vietnamese in-house data and recognized on HYKIST dataset.
    }
\label{table:L2_comb}
\end{table}

Due to time constraint and project requirement, we only tune L2 regularization values in favor of HYKIST data performance.
By using grid-search technique for right L2 value selection, we are able to reduce \glspl{WER} of multiple pretraining schedules as shown in Table \ref{table:L2_comb}.
Notable results are seen on raw waveform from scratch training, where \glspl{WER} reduce from 33.0\% and 38.8\% (only \glsxtrshort{IF Loss}) to 31.4\% and 36.4\% (combination with L2 regularization) on dev and test set respectively, that makes \glsxtrshort{WERR} 5.5\% in average.

We stick with default parameters for \glsxtrshort{ICE Loss} and \glsxtrshort{IF Loss} because \glsxtrshort{WER}s do not fluctuate significantly when choosing other parameters.
However, the right parameters for L2 regularization are chosen based on grid-search strategy and different parameters make the \glsxtrshort{WER}s vary greatly.
Therefore, we recommend the use of L2 regularization should be the last regularization effort in the entire regularization pipeline due to its higher sentitivity to \glsxtrshort{WER}s compared to \glsxtrshort{ICE Loss} and \glsxtrshort{IF Loss}.

    %\section{Literature comparison}

%\input{tables/literature_compare}

%In this part we evaluate our best models compared to \cite{Duy_Khanh_Finetune_Wav2vec_2_0_2022}'s model as shown in Table \ref{literature_compare}. 
%Although our fine-tuning data is less diverse than their combination of spontaneous-reading VLSP, CommonVoice, VIVOS datasets; and our \glsxtrshort{LM} is generated by far less text data (2GB text data vs. our 500M words), we still achieved better performance on HYKIST, e.g. \glsxtrshort{WERR} 10.8\% and 13.0\% in average for monolingual and multilingual models respectively. 
    
    \chapter{Conclusion}
\label{ch: Conclusion}

\section{Overall results}

In this thesis, we describe our efforts to develop HYKIST-related \glsxtrshort{ASR} systems for conversational telephone speech in the medical field for Vietnamese language.

Firstly, we use various acoustic encoder topologies to present supervised-only baselines while deploying the hybrid \glsxtrshort{HMM} framework.

Secondly, we use unsupervised \glsxtrshort{Wav2vec 2.0} pretraining to improve system performance and analyze the effects of pretraining data on performance.
The experimental findings demonstrate that this is especially effective when diverse pretraining data is used, e.g. data on multiple domains, multi-speaker data, augmented data...
Also, multilingual pretraining does not always outperform monolingual pretraining.
It is also shown that cost-effective model development is possible by utilizing the \glsxtrshort{XLSR-53} model, which is freely available.
We then compare with the baselines and show that the \glsxtrshort{Wav2vec 2.0} unsupervised pretraining does not always outperform the \glsxtrshort{Transformer} supervised-only approach, especially when the pretrained data is not diverse enough.

Thirdly, our best method to further improve the accuracy is using continued pretraining approach, where we pretrain multiple 8kHz datasets using parameters initialized by the 16kHz multilingual \glsxtrshort{XLSR-53} model.
We show that continued pretraining is beneficial for both the monolingual and the multilingual scenario.
However, the continued pretraining on less diverse data benefits more than the diverse data.

Fourthly, we compare the performance of \glsxtrshort{Wav2vec 2.0} encoders and recommend the \textit{Base} architecture instead of \textit{Large}\textsubscript{1-8} for the sake of both accuracy and inference performance.
We also recommend the use of Kaiming Initialization for better accuracy of \glsxtrshort{Wav2vec 2.0} architecture, instead of Xavier Initialization.

Finally, we apply and analyze the use of intermediate loss - \glsxtrfull{ICE Loss} and \glsxtrfull{IF Loss} - to make \glsxtrshort{Wav2vec 2.0} more robust for all recognition domains.
We prove that \glsxtrshort{IF Loss} works better than \glsxtrshort{ICE Loss} in all data domains.
In addition, for the small out-of-domain recognition \glsxtrshort{IF Loss} works well but for the large out-of-domain recognition it should only be applied on less diverse pretrained data.
In order to further improvement of accuracy, we integrate \glsxtrshort{IF Loss} with On-off Regularization and L2 Regularization.

\section{Future work}

During the work of this thesis, we have discovered some promising directions which are planned for future work.
First, section \ref{sec: unsupervised_pretraining} shows that the system performance benefits from the unsupervised pretraining on diverse data but pretraining on the in-domain data, medical speech data in other words, is not compared yet.
Second, we show that the data augmentation in pretraining stage is effective. However, such data augmentation for finetuning is not investigated yet.
Third, due to time constraint, the effectiveness of On-off Regularization for different pretraining schedules is not studied. 
This leads to the question if sequence discriminative training \cite{gibson2006hypothesis}, which also uses learning rate reset, works well with \glsxtrshort{Wav2vec 2.0}.
Finally, Wav2vec 2.0 - Conformer \cite{wav2vec2_conformer} has been popular lately. 
However, its effectiveness on Vietnamese has not been investigated yet.

    \appendix

    \printbibliography[heading=bibnumbered,
    title={Bibliography}]

@inproceedings{panayotov2015librispeech,
  title={Librispeech: an asr corpus based on public domain audio books},
  author={Panayotov, Vassil and Chen, Guoguo and Povey, Daniel and Khudanpur, Sanjeev},
  booktitle={2015 IEEE international conference on acoustics, speech and signal processing (ICASSP)},
  pages={5206--5210},
  year={2015},
  organization={IEEE}
}

@article{xlsr53,
  title={Unsupervised cross-lingual representation learning for speech recognition},
  author={Conneau, Alexis and Baevski, Alexei and Collobert, Ronan and Mohamed, Abdelrahman and Auli, Michael},
  journal={arXiv preprint arXiv:2006.13979},
  year={2020}
}

@article{wav2vec2,
  title={wav2vec 2.0: A framework for self-supervised learning of speech representations},
  author={Baevski, Alexei and Zhou, Yuhao and Mohamed, Abdelrahman and Auli, Michael},
  journal={Advances in Neural Information Processing Systems},
  volume={33},
  pages={12449--12460},
  year={2020}
}

@book{ethnologue,
  title = {Ethnologue: Languages of the World},
  year = {2009},
  researchr = {https://researchr.org/publication/ethnologue},
  cites = {0},
  citedby = {0},
  edition = {Sixteenth},
  editor = {M. Paul Lewis},
  address = {Dallas, TX, USA},
  publisher = {SIL International},
}

@inproceedings{BERT,
  title={BERT: Pre-training of Deep Bidirectional Transformers for Language Understanding},
  author={Kenton, Jacob Devlin Ming-Wei Chang and Toutanova, Lee Kristina},
  booktitle={Proceedings of NAACL-HLT},
  pages={4171--4186},
  year={2019}
}

@article{Transformer,
  title={Attention is all you need},
  author={Vaswani, Ashish and Shazeer, Noam and Parmar, Niki and Uszkoreit, Jakob and Jones, Llion and Gomez, Aidan N and Kaiser, {\L}ukasz and Polosukhin, Illia},
  journal={Advances in neural information processing systems},
  volume={30},
  year={2017}
}

@inproceedings{RASR-hybrid_vs_attention,
	title={{RWTH} {ASR} Systems for {LibriSpeech}: Hybrid vs {Attention}},
	author={Christoph L\"uscher and Eugen Beck and Kazuki Irie and Markus Kitza and Wilfried Michel and Albert Zeyer and Ralf Schl\"uter and Hermann Ney},
	booktitle=confInterspeech,
	address={Graz, Austria},
	month=sep,
	year={2019}
}

@inproceedings{BABEL_dataset,
  author = {Gales, Mark J. F. and Knill, Kate M. and Ragni, Anton and Rath, Shakti P.},
  booktitle = {SLTU},
  crossref = {conf/sltu/2014},
  ee = {http://www.isca-speech.org/archive/sltu_2014/sl14_016.html},
  pages = {16-23},
  publisher = {ISCA},
  title = {Speech recognition and keyword spotting for low-resource languages: Babel project research at CUED.},
  url = {http://dblp.uni-trier.de/db/conf/sltu/sltu2014.html#GalesKRR14},
  year = 2014
}

@inproceedings{CommonVoice_dataset,
    title = "Common Voice: A Massively-Multilingual Speech Corpus",
    author = "Ardila, Rosana  and
      Branson, Megan  and
      Davis, Kelly  and
      Kohler, Michael  and
      Meyer, Josh  and
      Henretty, Michael  and
      Morais, Reuben  and
      Saunders, Lindsay  and
      Tyers, Francis  and
      Weber, Gregor",
    booktitle = "Proceedings of the 12th Language Resources and Evaluation Conference",
    month = may,
    year = "2020",
    address = "Marseille, France",
    publisher = "European Language Resources Association",
    url = "https://aclanthology.org/2020.lrec-1.520",
    pages = "4218--4222",
    language = "English",
    ISBN = "979-10-95546-34-4",
}

@inproceedings{multiling_librispeech,
  title={MLS: A Large-Scale Multilingual Dataset for Speech Research},
  author={Pratap, Vineel and Xu, Qiantong and Sriram, Anuroop and Synnaeve, Gabriel and Collobert, Ronan},
  booktitle={INTERSPEECH},
  year={2020}
}

@article{facebook2020hybrid,
  title={Fast, Simpler and More Accurate Hybrid {ASR} Systems Using Wordpieces},
  author={Zhang, Frank and Wang, Yongqiang and Zhang, Xiaohui and Liu, Chunxi and Saraf, Yatharth and Zweig, Geoffrey},
  journal={arXiv preprint arXiv:2005.09150},
  year={2020}
}

@article{google2020conformer,
  title={Conformer: Convolution-augmented transformer for speech recognition},
  author={Gulati, Anmol and Qin, James and Chiu, Chung-Cheng and Parmar, Niki and Zhang, Yu and Yu, Jiahui and Han, Wei and Wang, Shibo and Zhang, Zhengdong and Wu, Yonghui and others},
  journal={arXiv preprint arXiv:2005.08100},
  year={2020}
}

@inproceedings{zeineldeen2022conformer,
  title={Conformer-based hybrid {ASR} system for Switchboard dataset},
  author={Zeineldeen, Mohammad and Xu, Jingjing and L{\"u}scher, Christoph and Michel, Wilfried and Gerstenberger, Alexander and Schl{\"u}ter, Ralf and Ney, Hermann},
  booktitle=confICASSP,
  pages={7437--7441},
  year={2022},
  organization={IEEE}
}

@inproceedings{vieting2021waveform,
  title={On Architectures and Training for Raw Waveform Feature Extraction in {ASR}},
  author={Vieting, Peter and L{\"u}scher, Christoph and Michel, Wilfried and Schl{\"u}ter, Ralf and Ney, Hermann},
  booktitle=confASRU,
  pages={267--274},
  year={2021},
  organization={IEEE}
}

@inproceedings{facebook2020dejavu,
  title={Deja-vu: Double feature presentation and iterated loss in deep transformer networks},
  author={Tjandra, Andros and Liu, Chunxi and Zhang, Frank and Zhang, Xiaohui and Wang, Yongqiang and Synnaeve, Gabriel and Nakamura, Satoshi and Zweig, Geoffrey},
  booktitle=confICASSP,
  pages={6899--6903},
  year={2020},
  organization={IEEE}
}

@article{mohamed2022representation_review,
  title={Self-Supervised Speech Representation Learning: A Review},
  author={Mohamed, Abdelrahman and Lee, Hung-yi and Borgholt, Lasse and Havtorn, Jakob D and Edin, Joakim and Igel, Christian and Kirchhoff, Katrin and Li, Shang-Wen and Livescu, Karen and Maal{\o}e, Lars and others},
  journal={arXiv preprint arXiv:2205.10643},
  year={2022}
}

@article{deepmind2020cpc,
  title={Learning robust and multilingual speech representations},
  author={Kawakami, Kazuya and Wang, Luyu and Dyer, Chris and Blunsom, Phil and Oord, Aaron van den},
  journal={arXiv preprint arXiv:2001.11128},
  volume={},
  year={2020}
}

@inproceedings{facebook2019wav2vec,
  title={wav2vec: Unsupervised Pre-Training for Speech Recognition},
  author={Schneider, Steffen and Baevski, Alexei and Collobert, Ronan and Auli, Michael},
  booktitle=confInterspeech,
  pages={3465--3469},
  address={Graz, Austria},
  month=sep,
  year={2019}
}

@inproceedings{facebook2021hubert,
  title={{HuBERT}: How much can a bad teacher benefit {ASR} pre-training?},
  author={Hsu, Wei-Ning and Tsai, Yao-Hung Hubert and Bolte, Benjamin and Salakhutdinov, Ruslan and Mohamed, Abdelrahman},
  booktitle=confICASSP,
  pages={6533--6537},
  year={2021},
  organization={IEEE}
}

@article{facebook2022wav2vecaug,
  title={Wav2Vec-Aug: Improved self-supervised training with limited data},
  author={Sriram, Anuroop and Auli, Michael and Baevski, Alexei},
  journal={arXiv preprint arXiv:2206.13654},
  year={2022}
}

@inproceedings{livescu2021wav2vec_analysis,
  title={Layer-wise analysis of a self-supervised speech representation model},
  author={Pasad, Ankita and Chou, Ju-Chieh and Livescu, Karen},
  booktitle=confASRU,
  pages={914--921},
  year={2021},
  organization={IEEE}
}

@inproceedings{robust_wav2vec2,
  title={Robust wav2vec 2.0: Analyzing Domain Shift in Self-Supervised Pre-Training},
  author={Wei-Ning Hsu and Anuroop Sriram and Alexei Baevski and Tatiana Likhomanenko and Qiantong Xu and Vineel Pratap and Jacob Kahn and Ann Lee and Ronan Collobert and Gabriel Synnaeve and Michael Auli},
  booktitle={Interspeech},
  year={2021}
}

@inproceedings{microsoft2021unispeech,
  title={Unispeech: Unified speech representation learning with labeled and unlabeled data},
  author={Wang, Chengyi and Wu, Yu and Qian, Yao and Kumatani, Kenichi and Liu, Shujie and Wei, Furu and Zeng, Michael and Huang, Xuedong},
  booktitle={International Conference on Machine Learning},
  pages={10937--10947},
  year={2021},
  organization={PMLR}
}

@article{zhang2021xlst,
  title={{XLST}: Cross-lingual self-training to learn multilingual representation for low resource speech recognition},
  author={Zhang, Zi-Qiang and Song, Yan and Wu, Ming-Hui and Fang, Xin and Dai, Li-Rong},
  journal={arXiv preprint arXiv:2103.08207},
  year={2021}
}

@inproceedings{google2022just,
  title={Joint unsupervised and supervised training for multilingual {ASR}},
  author={Bai, Junwen and Li, Bo and Zhang, Yu and Bapna, Ankur and Siddhartha, Nikhil and Sim, Khe Chai and Sainath, Tara N},
  booktitle=confICASSP,
  pages={6402--6406},
  year={2022},
  organization={IEEE}
}

@inproceedings{tuske2014multilingual,
  author={T{\"u}ske, Zolt{\'a}n and Golik, Pavel and Nolden, David and Schl{\"u}ter, Ralf and Ney, Hermann},
  title={Data Augmentation, Feature Combination, and Multilingual Neural Networks to Improve {ASR} and {KWS} Performance for Low-resource Languages},
  booktitle=confInterspeech,
  year=2014,
  pages={1420-1424},
  address={Singapore},
  month=sep
}

@InProceedings{edwards2017medicalspeech,
author="Edwards, Erik
and Salloum, Wael
and Finley, Greg P.
and Fone, James
and Cardiff, Greg
and Miller, Mark
and Suendermann-Oeft, David",
editor="Karpov, Alexey
and Potapova, Rodmonga
and Mporas, Iosif",
title="Medical Speech Recognition: Reaching Parity with Humans",
booktitle="Speech and Computer",
year="2017",
publisher="Springer International Publishing",
address="Cham",
pages="512--524",
isbn="978-3-319-66429-3"
}

@inproceedings{chiu2018medconv,
    author={Chung-Cheng Chiu and Anshuman Tripathi and Katherine Chou and Chris Co and Navdeep Jaitly and Diana Jaunzeikare and Anjuli Kannan and Patrick Nguyen and Hasim Sak and Ananth Sankar and Justin Tansuwan and Nathan Wan and Yonghui Wu and Xuedong Zhang},
    title={{Speech Recognition for Medical Conversations}},
    year=2018,
    booktitle={Proc. Interspeech 2018},
    pages={2972--2976},
    doi={10.21437/Interspeech.2018-40}
}

@inproceedings{kar2021operation,
    author={Kar, Snigdhaswin and Mishra, Prabodh and Lin, Ju and Woo, Min-Jae and Deas, Nicholas and Linduff, Caleb and Niu, Sufeng and Yang, Yuzhe and McClendon, Jerome and Smith, D. Hudson and Smith, Melissa C. and Gimbel, Ronald W. and Wang, Kuang-Ching},
    booktitle={2021 International Joint Conference on Neural Networks (IJCNN)},
    title={Systematic Evaluation and Enhancement of Speech Recognition in Operational Medical Environments},
    year={2021},
    volume={},
    number={},
    pages={1-8},
    doi={10.1109/IJCNN52387.2021.9533607}
}

@inproceedings{sakti2014towards,
    title = "Towards Multilingual Conversations in the Medical Domain: Development of Multilingual Medical Data and A Network-based {ASR} System",
    author = "Sakti, Sakriani  and
              Kubo, Keigo  and
              Matsumiya, Sho  and
              Neubig, Graham  and
              Toda, Tomoki  and
              Nakamura, Satoshi  and
              Adachi, Fumihiro  and
              Isotani, Ryosuke",
    booktitle = "Proceedings of the Ninth International Conference on Language Resources and Evaluation ({LREC}'14)",
    month = may,
    year = "2014",
    address = "Reykjavik, Iceland",
    publisher = "European Language Resources Association (ELRA)",
    url = "http://www.lrec-conf.org/proceedings/lrec2014/pdf/709_Paper.pdf",
    pages = "2639--2643",
}

@inproceedings{mani2020towardsmedical,
    title={Towards Understanding ASR Error Correction for Medical Conversations},
    author={Anirudh Mani and Shruti Palaskar and Sandeep Konam},
    booktitle={NLPMC},
    year={2020}
}

@misc{layer_normalization,
  doi = {10.48550/ARXIV.1607.06450},
  
  url = {https://arxiv.org/abs/1607.06450},
  
  author = {Ba, Jimmy Lei and Kiros, Jamie Ryan and Hinton, Geoffrey E.},
  
  title = {Layer Normalization},
  
  publisher = {arXiv},
  
  year = {2016},
  
  copyright = {arXiv.org perpetual, non-exclusive license}
}

@article{gelu,
  author    = {Dan Hendrycks and
               Kevin Gimpel},
  title     = {Bridging Nonlinearities and Stochastic Regularizers with Gaussian
               Error Linear Units},
  journal   = {CoRR},
  volume    = {abs/1606.08415},
  year      = {2016},
  url       = {http://arxiv.org/abs/1606.08415},
  eprinttype = {arXiv},
  eprint    = {1606.08415},
  bibsource = {dblp computer science bibliography, https://dblp.org}
}

@MISC{wav2vec2_towardsdatascience,
   author =       {Łukasz Sus},
   title =        {Wav2Vec 2.0: A Framework for Self-Supervised Learning of Speech Representations - Model for speech recognition explained
   },
   editor =       {towardsdatascience.com},
   month =        {June},
   year =         {2021},
   url = {https://towardsdatascience.com/wav2vec-2-0-a-framework-for-self-supervised-learning-of-speech-representations-7d3728688cae},
}

@article{product_quantization,
  title={Product Quantization for Nearest Neighbor Search},
  author={Herv{\'e} J{\'e}gou and Matthijs Douze and Cordelia Schmid},
  journal={IEEE Transactions on Pattern Analysis and Machine Intelligence},
  year={2011},
  volume={33},
  pages={117-128}
}

@article{lample2019cross,
  title={Cross-lingual language model pretraining},
  author={Lample, Guillaume and Conneau, Alexis},
  journal={arXiv preprint arXiv:1901.07291},
  year={2019}
}

@INPROCEEDINGS{kneser_ney_lm,
  author={Kneser, R. and Ney, H.},
  booktitle={1995 International Conference on Acoustics, Speech, and Signal Processing}, 
  title={Improved backing-off for M-gram language modeling}, 
  year={1995},
  volume={1},
  number={},
  pages={181-184 vol.1},
  doi={10.1109/ICASSP.1995.479394}
}

@article{beck2019lstm,
  title={Lstm language models for lvcsr in first-pass decoding and lattice-rescoring},
  author={Beck, Eugen and Zhou, Wei and Schl{\"u}ter, Ralf and Ney, Hermann},
  journal={arXiv preprint arXiv:1907.01030},
  year={2019}
}

@article{bisani2008g2p,
    title = {Joint-sequence models for grapheme-to-phoneme conversion},
    journal = {Speech Communication},
    volume = {50},
    number = {5},
    pages = {434-451},
    year = {2008},
    issn = {0167-6393},
    doi = {https://doi.org/10.1016/j.specom.2008.01.002},
    url = {https://www.sciencedirect.com/science/article/pii/S0167639308000046},
    author = {Maximilian Bisani and Hermann Ney}
}

@INPROCEEDINGS{stolcke2002srilm,
    author = {Andreas Stolcke},
    title = {SRILM -- An extensible language modeling toolkit},
    booktitle = {IN PROCEEDINGS OF THE 7TH INTERNATIONAL CONFERENCE ON SPOKEN LANGUAGE PROCESSING (ICSLP 2002)},
    year = {2002},
    pages = {901--904},
    publisher = {},
}

@inproceedings{doetsch2016returnn,
  title={{RETURNN}: the {RWTH} extensible training framework for universal recurrent neural networks},
  author={Doetsch, Patrick and Zeyer, Albert and Voigtlaender, Paul and Kulikov, Ilya and Schl{\"u}ter, Ralf and Ney, Hermann},
  booktitle =confICASSP,
  address={New Orleans, {LA}, {USA}},
  year={2017}
}

@inproceedings{facebook2019fairseq,
  title={fairseq: A Fast, Extensible Toolkit for Sequence Modeling},
  author={Ott, Myle and Edunov, Sergey and Baevski, Alexei and Fan, Angela and Gross, Sam and Ng, Nathan and Grangier, David and Auli, Michael},
  booktitle={Proceedings of the 2019 Conference of the North American Chapter of the Association for Computational Linguistics (Demonstrations)},
  pages={48--53},
  year={2019}
}

@InProceedings {rybach2011rasr,
    author= {Rybach, David and Hahn, Stefan and Lehnen, Patrick and Nolden, David and Sundermeyer, Martin and T{\"u}ske, Zolt{\'a}n and Wiesler, Simon and Schl{\"u}ter, Ralf and Ney, Hermann},
    title= {{RASR} - The {RWTH Aachen University} Open Source Speech Recognition Toolkit},
    booktitle=confASRU,
    year= 2011,
    address= {Waikoloa, HI, USA},
    month= dec,
    booktitlelink= {http://www.asru2011.org}
}

@InProceedings { schlueter:icassp07,
	author= {Schl{\"u}ter, Ralf and Bezrukov, Ilja and Wagner, Hermann and Ney, Hermann},	
	title= {Gammatone Features and Feature Combination for Large Vocabulary Speech Recognition},	
	booktitle= confICASSP,	
	year= 2007,	
	pages= {649--652},	
	address= {Honolulu, HI, USA},	
	month= apr,	
}

@article{dropout,
  author  = {Nitish Srivastava and Geoffrey Hinton and Alex Krizhevsky and Ilya Sutskever and Ruslan Salakhutdinov},
  title   = {Dropout: A Simple Way to Prevent Neural Networks from Overfitting},
  journal = {Journal of Machine Learning Research},
  year    = {2014},
  volume  = {15},
  number  = {56},
  pages   = {1929--1958},
  url     = {http://jmlr.org/papers/v15/srivastava14a.html}
}

@article{park2019specaugment,
  title={Specaugment: A simple data augmentation method for automatic speech recognition},
  author={Park, Daniel S and Chan, William and Zhang, Yu and Chiu, Chung-Cheng and Zoph, Barret and Cubuk, Ekin D and Le, Quoc V},
  journal={arXiv preprint arXiv:1904.08779},
  year={2019}
}

@article{hochreiter1997long,
  title={Long short-term memory},
  author={Hochreiter, Sepp and Schmidhuber, J{\"u}rgen},
  journal={Neural computation},
  volume={9},
  number={8},
  pages={1735--1780},
  year={1997}
}

@misc{zeineldeen2022robustconformer,
  doi = {10.48550/ARXIV.2206.12955},
  
  url = {https://arxiv.org/abs/2206.12955},
  
  author = {Zeineldeen, Mohammad and Xu, Jingjing and Lüscher, Christoph and Schlüter, Ralf and Ney, Hermann},
  
  title = {Improving the Training Recipe for a Robust Conformer-based Hybrid Model},
  
  publisher = {arXiv},
  
  year = {2022},
  
  copyright = {arXiv.org perpetual, non-exclusive license}
}

@inproceedings{focal_loss,
  title={Focal loss for dense object detection},
  author={Lin, Tsung-Yi and Goyal, Priya and Girshick, Ross and He, Kaiming and Doll{\'a}r, Piotr},
  booktitle={Proceedings of the IEEE international conference on computer vision},
  pages={2980--2988},
  year={2017}
}

@inproceedings{L2_regularization,
    author = {Krogh, Anders and Hertz, John},
    booktitle = {Advances in Neural Information Processing Systems},
    editor = {J. Moody and S. Hanson and R.P. Lippmann},
    pages = {},
    publisher = {Morgan-Kaufmann},
    title = {A Simple Weight Decay Can Improve Generalization},
    url = {https://proceedings.neurips.cc/paper/1991/file/8eefcfdf5990e441f0fb6f3fad709e21-Paper.pdf},
    volume = {4},
    year = {1991}
}

@inproceedings{speed_perturbation,
  title={Audio augmentation for speech recognition},
  author={Tom Ko and Vijayaditya Peddinti and Daniel Povey and Sanjeev Khudanpur},
  booktitle={INTERSPEECH},
  year={2015}
}

@article{pitch_perturbation,
  title={Data Augmentation For Children's Speech Recognition--The" Ethiopian" System For The SLT 2021 Children Speech Recognition Challenge},
  author={Chen, Guoguo and Na, Xingyu and Wang, Yongqing and Yan, Zhiyong and Zhang, Junbo and Ma, Sifan and Wang, Yujun},
  journal={arXiv preprint arXiv:2011.04547},
  year={2020}
}

@article{reverb_perturbation,
  title={JHU ASpIRE system: Robust LVCSR with TDNNS, iVector adaptation and RNN-LMS},
  author={Vijayaditya Peddinti and Guoguo Chen and Vimal Manohar and Tom Ko and Daniel Povey and Sanjeev Khudanpur},
  journal={2015 IEEE Workshop on Automatic Speech Recognition and Understanding (ASRU)},
  year={2015},
  pages={539-546}
}

@article{irie2019language,
  title={Language Modeling with Deep Transformers},
  author={Irie, Kazuki and Zeyer, Albert and Schl{\"u}ter, Ralf and Ney, Hermann},
  journal={Proc. Interspeech 2019},
  pages={3905--3909},
  year={2019}
}

@inproceedings{lee2021intermediate,
  title={Intermediate loss regularization for ctc-based speech recognition},
  author={Lee, Jaesong and Watanabe, Shinji},
  booktitle={ICASSP 2021-2021 IEEE International Conference on Acoustics, Speech and Signal Processing (ICASSP)},
  pages={6224--6228},
  year={2021},
  organization={IEEE}
}

@inproceedings{ardila2020commonvoice,
  title={Common Voice: A Massively-Multilingual Speech Corpus},
  author={Ardila, Rosana and Branson, Megan and Davis, Kelly and Kohler, Michael and Meyer, Josh and Henretty, Michael and Morais, Reuben and Saunders, Lindsay and Tyers, Francis and Weber, Gregor},
  booktitle={Proceedings of the 12th Language Resources and Evaluation Conference},
  pages={4218--4222},
  year={2020}
}

@inproceedings{vivos_dataset,
  title={A non-expert Kaldi recipe for Vietnamese speech recognition system},
  author={Luong, Hieu-Thi and Vu, Hai-Quan},
  booktitle={Proceedings of the Third International Workshop on Worldwide Language Service Infrastructure and Second Workshop on Open Infrastructures and Analysis Frameworks for Human Language Technologies (WLSI/OIAF4HLT2016)},
  pages={51--55},
  year={2016}
}

@inproceedings{wav2vec2_conformer,
  author={Edwin G. Ng and Chung-Cheng Chiu and Yu Zhang and William Chan},
  title={{Pushing the Limits of Non-Autoregressive Speech Recognition}},
  year=2021,
  booktitle={Proc. Interspeech 2021},
  pages={3725--3729},
  doi={10.21437/Interspeech.2021-337}
}

@inproceedings{gibson2006hypothesis,
  title={Hypothesis spaces for minimum Bayes risk training in large vocabulary speech recognition.},
  author={Gibson, Matthew and Hain, Thomas},
  organization={Citeseer}
}

@misc{Aertsen_Olders_Johannesma_1981, title={Spectro-temporal receptive fields of auditory neurons in the grassfrog}, volume={39}, url={http://dx.doi.org/10.1007/BF00342772}, DOI={10.1007/bf00342772}, number={3}, journal={Biological Cybernetics}, publisher={Springer Science and Business Media LLC}, author={Aertsen, A. M. H. J. and Olders, J. H. J. and Johannesma, P. I. M.}, year={1981}, pages={195–209}, language={en} }

@misc{Greenwood_1990, title={A cochlear frequency‐position function for several species—29 years later}, volume={87}, url={http://dx.doi.org/10.1121/1.399052}, DOI={10.1121/1.399052}, number={6}, journal={The Journal of the Acoustical Society of America}, publisher={Acoustical Society of America (ASA)}, author={Greenwood, Donald D.}, year={1990}, month={Jun}, pages={2592–2605}, language={en} }

@book{Rao_KE_2017, title={Speech Recognition Using Articulatory and Excitation Source Features}, url={http://dx.doi.org/10.1007/978-3-319-49220-9}, DOI={10.1007/978-3-319-49220-9}, journal={SpringerBriefs in Electrical and Computer Engineering}, publisher={Springer International Publishing}, author={Rao, K. Sreenivasa and K E, Manjunath}, year={2017} }

@article{zhang2021dive,
  title={Dive into deep learning},
  author={Zhang, Aston and Lipton, Zachary C and Li, Mu and Smola, Alexander J},
  journal={arXiv preprint arXiv:2106.11342},
  year={2021}
}

@inproceedings{relu,
author = {Nair, Vinod and Hinton, Geoffrey E.},
title = {Rectified Linear Units Improve Restricted Boltzmann Machines},
year = {2010},
isbn = {9781605589077},
publisher = {Omnipress},
address = {Madison, WI, USA},
booktitle = {Proceedings of the 27th International Conference on International Conference on Machine Learning},
pages = {807–814},
numpages = {8},
location = {Haifa, Israel},
series = {ICML'10}
}

@misc{amidi2018deep,
  title={Deep Learning Cheatsheet},
  author={Amidi, Afshine and Amidi, Shervine},
  year={2018},
  publisher={CS}
}

@inproceedings{kingma2015adam,
  title={Adam: A Method for Stochastic Optimization},
  author={Kingma, Diederik P and Ba, Jimmy},
  booktitle={ICLR (Poster)},
  year={2015}
}

@InProceedings{xavier_init, title = 	 {Understanding the difficulty of training deep feedforward neural networks}, author = 	 {Glorot, Xavier and Bengio, Yoshua}, booktitle = 	 {Proceedings of the Thirteenth International Conference on Artificial Intelligence and Statistics}, pages = 	 {249--256}, year = 	 {2010}, editor = 	 {Teh, Yee Whye and Titterington, Mike}, volume = 	 {9}, series = 	 {Proceedings of Machine Learning Research}, address = 	 {Chia Laguna Resort, Sardinia, Italy}, month = 	 {13--15 May}, publisher =    {PMLR}, url = 	 {https://proceedings.mlr.press/v9/glorot10a.html} }

@article{lstm1997,
author = {Hochreiter, Sepp and Schmidhuber, J\"{u}rgen},
title = {Long Short-Term Memory},
year = {1997},
issue_date = {November 15, 1997},
publisher = {MIT Press},
address = {Cambridge, MA, USA},
volume = {9},
number = {8},
issn = {0899-7667},
url = {https://doi.org/10.1162/neco.1997.9.8.1735},
doi = {10.1162/neco.1997.9.8.1735},
journal = {Neural Comput.},
month = {nov},
pages = {1735–1780},
numpages = {46}
}

@ARTICLE{brnn1997,  author={Schuster, M. and Paliwal, K.K.},  journal={IEEE Transactions on Signal Processing},   title={Bidirectional recurrent neural networks},   year={1997},  volume={45},  number={11},  pages={2673-2681},  doi={10.1109/78.650093}}

@INPROCEEDINGS{DeepResidualLearning,  author={He, Kaiming and Zhang, Xiangyu and Ren, Shaoqing and Sun, Jian},  booktitle={2016 IEEE Conference on Computer Vision and Pattern Recognition (CVPR)},   title={Deep Residual Learning for Image Recognition},   year={2016},  volume={},  number={},  pages={770-778},  doi={10.1109/CVPR.2016.90}}

@article{Baum1967AnIW,
  title={An inequality with applications to statistical estimation for probabilistic functions of Markov processes and to a model for ecology},
  author={Leonard E. Baum and John A. Eagon},
  journal={Bulletin of the American Mathematical Society},
  year={1967},
  volume={73},
  pages={360-363}
}

@INPROCEEDINGS{Beulen98automaticquestion,
  author = {K. Beulen and H. Ney},
    title = {Automatic Question Generation For Decision Tree Based State Tying},
    booktitle = {Proceedings of the IEEE Conference on Acoustics, Speech and Signal Processing},
    year = {1998},
    pages = {805--809}
}

@article{ortmanns1997word,
  title={A word graph algorithm for large vocabulary continuous speech recognition},
  author={Ortmanns, Stefan and Ney, Hermann and Aubert, Xavier},
  journal={Computer Speech \& Language},
  volume={11},
  number={1},
  pages={43--72},
  year={1997},
  publisher={Elsevier}
}

@article{dozat2016incorporating,
  title={Incorporating nesterov momentum into adam},
  author={Dozat, Timothy},
  year={2016}
}

@InProceedings{He_2015_ICCV,
author = {He, Kaiming and Zhang, Xiangyu and Ren, Shaoqing and Sun, Jian},
title = {Delving Deep into Rectifiers: Surpassing Human-Level Performance on ImageNet Classification},
booktitle = {Proceedings of the IEEE International Conference on Computer Vision (ICCV)},
month = {December},
year = {2015}
}

@inproceedings{glorot2010understanding,
  title={Understanding the difficulty of training deep feedforward neural networks},
  author={Glorot, Xavier and Bengio, Yoshua},
  booktitle={Proceedings of the thirteenth international conference on artificial intelligence and statistics},
  pages={249--256},
  year={2010},
  organization={JMLR Workshop and Conference Proceedings}
}

@Unpublished {luescher2022:hykist,
    author= {Lüscher, Christoph and Zeineldeen, Mohammad and Yang, Zijian and Vieting, Peter and Le-Duc, Khai and Wang, Weiyue and Schlüter, Ralf and Ney, Hermann},
    title= {Development of Hybrid ASR Systems for Low Resource Medical Domain Conversational Telephone Speech},
    note= {Baseline systems for Arabic, German, and Vietnamese for HYKIST project. Submitted to ICASSP 2023.},
    month= oct,
    year= 2022,
    url = {http://https://arxiv.org/abs/2210.13397}
}

    \glsaddall[types={symbolslist,glossaryx}]

    % This command would add *all** non-mentioned glossary entries (e.g. the acronyms)
    % \glsaddall
    % It's better to mention those in the text via 
    % \acrlong{BMS}, \acrshort{BMS}, and \acrfull{BMS}.
    % Because then the page numbers occur correctly in the list of acronyms
    \clearpage

    %\printglossary[type=symbolslist,style=symbunitlong,title=List of Symbols]
    \printglossary[type=acronym, title=List of Abbreviations and Glossaries]

    %\printglossary[type=glossaryx,style=glossaryxstyle,title=Glossary]
    %\printglossary[type=glossaryx, title=Glossary]
   
    \listoffigures
    \listoftables
    
\end{spacing}

\end{document}